\documentclass[runningheads]{llncs}
\usepackage[T1]{fontenc}
\usepackage{graphicx}
\usepackage{multirow}
\usepackage{longtable}
\usepackage{amsmath}
\usepackage{booktabs}
\usepackage{subcaption}

\usepackage{color}
\usepackage{hyperref}

\usepackage{cleveref}

\begin{document}
\title{ESPLoRA: Enhanced Spatial Precision with Low-Rank Adaption in Text-to-Image Diffusion Models for High-Definition Synthesis}

\titlerunning{ESPLoRA: Enhanced Spatial Precision in Diffusion Models}
\author{Andrea Rigo\inst{1} \and
Luca Stornaiuolo\inst{2} \and
Mauro Martino\inst{3} \and
Bruno Lepri\inst{4} \and
Nicu Sebe\inst{1}}

\authorrunning{A. Rigo et al.}

\institute{DISI, University of Trento, Trento, Italy \and
Toretei S.r.l., Roma, Italy \and
Visual AI Lab, MIT-IBM Watson AI Lab, Cambridge, Massachusetts, U.S.A. \and
Fondazione Bruno Kessler, Trento, Italy
}

\maketitle
\begin{abstract}
Diffusion models have revolutionized text-to-image (T2I) synthesis, producing high-quality, photorealistic images. However, they still struggle to properly render the spatial relationships described in text prompts. 
To address the lack of spatial information in T2I generations, existing methods typically use external network conditioning and predefined layouts, resulting in higher computational costs and reduced flexibility.
Our approach builds upon a curated dataset of spatially explicit prompts, meticulously extracted and synthesized from LAION-400M to ensure precise alignment between textual descriptions and spatial layouts. Alongside this dataset, we present ESPLoRA, a flexible fine-tuning framework based on Low-Rank Adaptation, specifically designed to enhance spatial consistency in generative models without increasing generation time or compromising the quality of the outputs.
In addition to ESPLoRA, we propose refined evaluation metrics grounded in geometric constraints, capturing 3D spatial relations such as \textit{in front of} or \textit{behind}. These metrics also expose spatial biases in T2I models which, even when not fully mitigated, can be strategically exploited by our TORE algorithm to further improve the spatial consistency of generated images. Our method outperforms CoMPaSS, the current baseline framework, on spatial consistency benchmarks.
\end{abstract}
\section{Introduction}
\label{sec:intro}

\begin{figure}[t]
    \centering
    \includegraphics[width=0.8\linewidth]{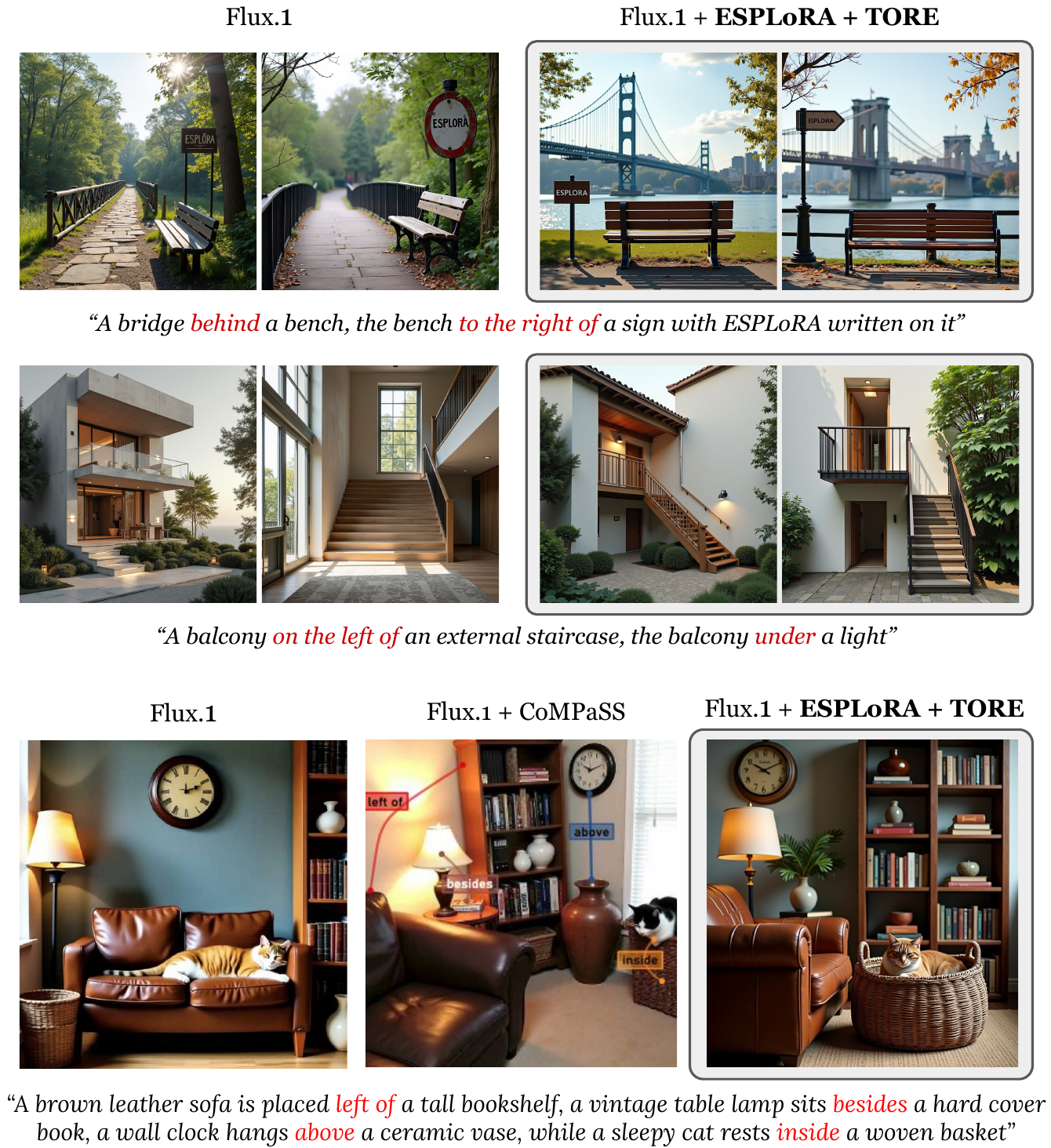}
    \caption{ESPLoRA enables existing T2I diffusion models to generate challenging spatial configurations, enhancing spatial capabilities without compromising output quality or increasing generation time.}
    \label{fig:base-vs-us}
    \vspace{-0.4cm}
\end{figure}

\begin{sloppypar}
Recent advancements in text-to-image generation, driven by diffusion-based models (e.g. Stable Diffusion \cite{rombach2022high}, Flux.1 \cite{flux2024}) and large-scale pretraining, have demonstrated remarkable performance in rendering photorealistic or stylistically consistent images from textual prompts \cite{dhariwal_diffusion_2021}. In this context, Low-Rank Adaptation (LoRA) \cite{hu2022lora} has proven to be highly efficient in the fine-tuning of models for learning diverse concepts, including objects, people, animals, and artistic styles. In particular, LoRA effectively captures and integrates new visual patterns while ensuring computational efficiency and preserving the knowledge of the pre-trained model. However, one of the major challenges still faced by these models is accurately representing spatial relationships among objects described in the prompt \cite{huang_t2i-compbench_2023,gokhale_benchmarking_2023,ghosh_geneval_2023,chatterjee_getting_2024,zhang_compass_2024}. This limitation is particularly pronounced in emerging AI-driven urban planning tools, where generating urban scenarios that adhere to specific policies and design constraints is crucial for realistic and functional city modeling. For instance, prompts such as “A balcony on the left of an external staircase, the balcony under a light” are often generated incorrectly due to the limited understanding of explicit positional constraints (e.g., “on the left of”, “under”, “in front of”).
Previous works \cite{chatterjee_getting_2024,zhang_compass_2024} show that this limitation is mainly caused by missing or ambiguous spatial descriptions in prompts of T2I training datasets.
Inspired by this, we propose a pipeline to detect spatial relationships in images and generate captions correctly aligned with them, building a dataset of spatially explicit and accurate text prompts and image pairs.
Then, by injecting specialized spatial knowledge into a pre-trained text-to-image model, we significantly improve its ability to follow spatial constraints in user prompts without compromising outputs quality.
We devise a pipeline to detect relationships on natural LAION \cite{schuhmann2021laion400mopendatasetclipfiltered} images by enforcing strict geometric constraints, extracting clear and non-ambiguous relationship-image pairs, which we use to generate spatially explicit prompts.
This results in a dataset of 5+ million relationships extracted from real-world urban scenarios.
We then generate images using these prompts and, by applying the same geometric constraints employed in our relationship extraction pipeline, we are able to select those that accurately reflect the spatial relationships expressed in the prompts, resulting in a new T2I dataset composed of spatially accurate prompts and synthetic images.
Finally, we fine-tune T2I models using LoRA on our proposed synthetic dataset and show that our approach can significantly improve the models capability of following the spatial relationships in the prompt.
Other works \cite{hu_ella_2024,feng_layoutgpt_2023} seek to exploit LLMs as reasoning engines to provide spatial signals to T2I models, which significantly increases the computational cost and time of the generation process.
In contrast, our approach, which we call ESPLoRA (Enhanced Spatial Precision with Low-Rank Adaption), achieves efficient performance by fine-tuning a lightweight adapter with negligible computational overhead. 
Additionally, we introduce a more precise benchmark based on spatial prompts and geometric constraints, enabling the evaluation of both single and multi relation prompts rather than focusing solely on one relationship.
Finally, we show that T2I models exhibit systematic biases when rendering specific spatial relationships (i.e., top, left, front), which persist even after fine-tuning. We propose TORE (Transforming Original Relations Effectively), an effective pre-processing step that leverages these biases to improve performance on the T2I-CompBench benchmark, with no additional computational cost. As we demonstrate in our results, naively applying LoRA to spatial understanding yields suboptimal results. Our key insight is that the combination of (i) strictly validated synthetic training data, (ii) geometry-grounded relationship extraction, and (iii) bias-aware prompt transformation creates a synergistic framework that significantly outperforms each component in isolation.
\end{sloppypar}

In summary, our contributions are:
\begin{itemize}
    \item A pipeline of models and a set of geometric constraints to detect 2D and 3D spatial relationships %, including 3D spatial relations in images such as \textit{in front of} or \textit{behind}.
    \item A new, open, high-resolution T2I dataset focused on spatial understanding % and urban scenarios.
    \item An adaptable fine-tuning framework, ESPLoRA, to inject spatial knowledge in T2I Diffusion Models with minimal computational overhead, based on synthetic images.
    \item An algorithm, TORE, that can exploit the model bias towards some relationships to improve its performance at spatial relationship rendering.
\end{itemize}

\begin{figure}[t]
    \centering
    \includegraphics[width=0.7\linewidth]{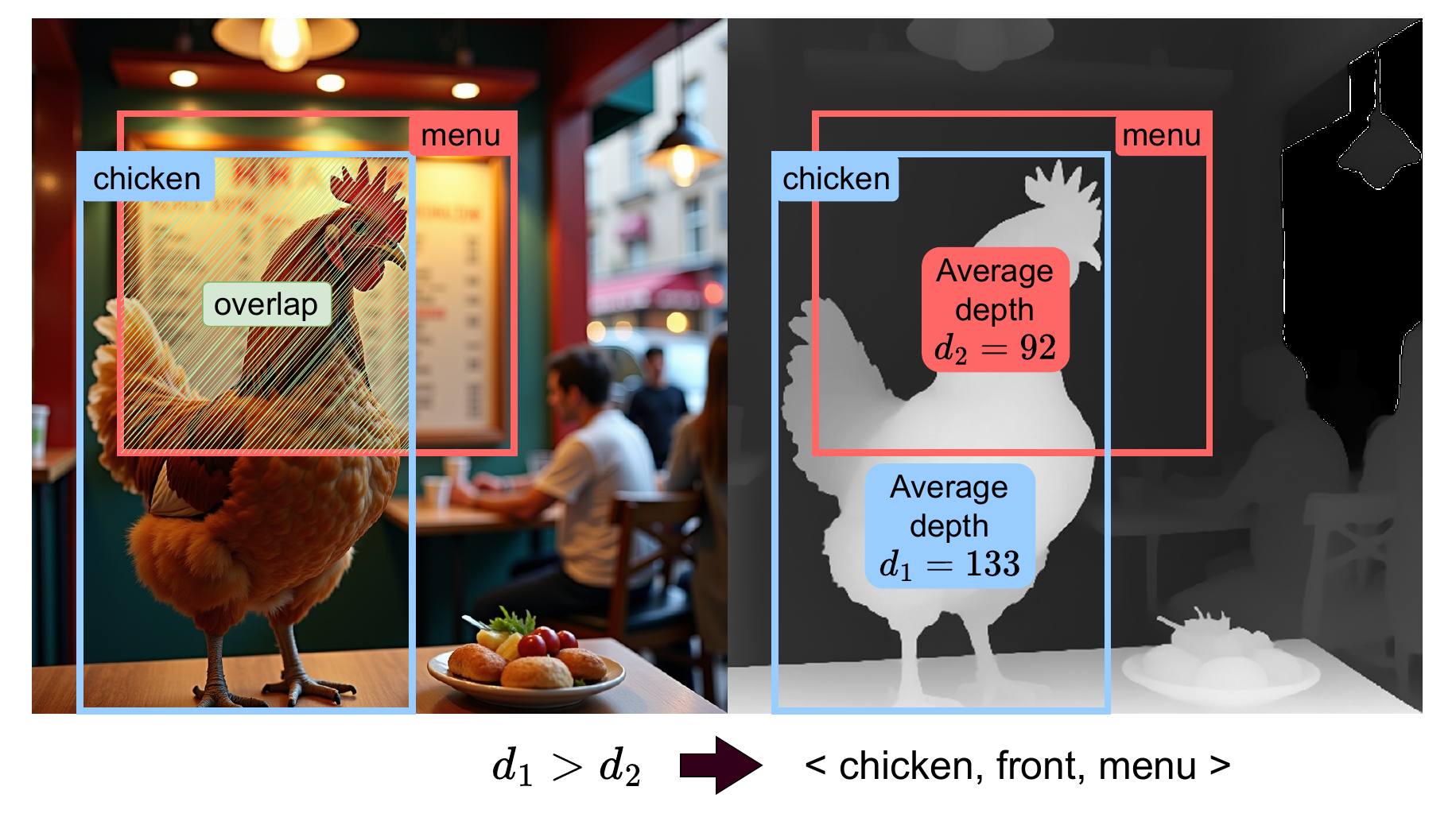}
    % \vspace{-0.4cm}    
    \caption{Example of a 3D relationship captured by our metric. On the left we check that the two objects overlap, on the right we compute the average depth of each object, and assign the correct 3D relationship accordingly.}
    \label{fig:3D-rel-example}
    \vspace{-0.4cm}
\end{figure}

\section{Related Work}
\label{sec:related-works}
\subsection{Controllable Text-to-Image Diffusion Models}
Recent research has focused on enhancing spatial control through training-based architectures and text-encoding strategies. Training-based approaches such as GLIGEN \cite{li_gligen_2023}, ReCo \cite{yang2022reco}, and ControlNet \cite{zhang_adding_2023} introduce explicit spatial conditioning into T2I models, while Uni-ControlNet \cite{zhao_uni-controlnet_2023} and T2I-Adapter \cite{mou_t2i-adapter_2023} propose lightweight modules to condition generation on reference signals. To bridge the gap between text and layout, several works leverage Large Language Models (LLMs). LayoutGPT \cite{feng_layoutgpt_2023} and LLM-Grounded Diffusion \cite{lian_llm-grounded_2023} utilize LLMs to generate spatial supervision signals, while others like ELLA \cite{hu_ella_2024} and ParaDiffusion \cite{wu_paragraph--image_2023} employ LLMs as robust text encoders. Furthermore, DiffusionGPT \cite{qin2024diffusiongpt} uses LLMs as reasoning engines to construct ad-hoc pipelines, and self-correcting frameworks \cite{wu_self-correcting_2023} iteratively refine images to match complex prompts.

\subsection{Spatial relationships in T2I models}
Beyond structural conditioning, other research addresses the inherent difficulty diffusion models face with spatial relationships and relational terminology.

\noindent Attention-based approaches such as \cite{chefer_attend-and-excite_2023,phung_grounded_2023}, enhance compositional generation by explicitly supervising attention-maps at inference time.
Some works use Large Language Models (LLMs) to generate spatial supervision signals to guide Diffusion Models. LayoutGPT \cite{feng_layoutgpt_2023} and LLM-Grounded Diffusion \cite{lian_llm-grounded_2023} generate layouts, and Control-GPT \cite{zhang_controllable_2023} generates TikZ layouts.
Other works like ELLA \cite{hu_ella_2024}, ParaDiffusion \cite{wu_paragraph--image_2023} and others \cite{koh_generating_2023,fu_guiding_2024,zhao2024bridging} use LLMs as text encoders.
DiffusionGPT \cite{qin2024diffusiongpt} uses an LLM as a reasoning engine to select some T2I models from a pool, constructing an ad-hoc pipeline. Self-correcting LLM-controlled Diffusion Models
\cite{wu_self-correcting_2023} instead implements a pipeline that can iteratively correct the generated image until it accurately matches the prompt.
Other works \cite{fu_guiding_2024}, like InstructPix2Pix \cite{brooks_instructpix2pix_2023}, take an orthogonal direction by improving prompt-image alignment via image editing.
A few benchmarks were devised to specifically evaluate the spatial consistency of Diffusion Models, such as VISOR \cite{gokhale_benchmarking_2023}, T2I-CompBench \cite{huang_t2i-compbench_2023} and GenEval \cite{ghosh_geneval_2023}.
More recently, promising works such as CoMPaSS \cite{zhang_compass_2024} and SPRIGHT \cite{chatterjee_getting_2024} demonstrated that the poor spatial relationships understanding in Diffusion Models is mainly caused by the absence or ambiguity of spatial terminology in prompts in T2I datasets. Both of them propose a curated dataset with synthetic captions where spatial relationships are properly expressed. CoMPaSS extracts unambiguous relationships from COCO \cite{lin2015microsoftcococommonobjects} using bounding boxes and some geometric rules and builds prompts from them, SPRIGHT uses LLaVA \cite{liu_visual_2023} to re-caption images from some large scale T2I datasets.
We position our work within these approaches, since LLM-based approaches substantially increase computational cost and generation time. Moreover, both adapter-based and attention-based methods typically rely on external supervision signals produced prior to generation in order to be effective, which reduces the flexibility of both input prompts and potential outputs.
In contrast, a LoRA-based approach achieves strong performance without the need for additional supervision, while introducing only minimal computational overhead and without any degradation in output quality.

\section{Methodology}
\label{sec:methodology}
Imprecise spatial relationships in large-scale T2I datasets are one of the leading reasons for poor spatial relationship understanding in Diffusion Models. This has been confirmed in two recent works \cite{chatterjee_getting_2024,zhang_compass_2024}.
Therefore, we devise a pipeline to extract clear object-relationships-object triplets from LAION-400M \cite{schuhmann2021laion400mopendatasetclipfiltered}, and use them to procedurally generate synthetic captions that accurately represent the spatial relationships in the image.
The relationships taken into consideration are the 2D \textit{right, left, top, bottom, next, between} and the 3D \textit{in front} and \textit{behind}.
We focus on complex urban scenarios and extract the urban context on each prompt, such as "street" or "city."

We created a novel ESPLoRA urban-realistic database using the proposed pipeline composed of the following stages:
\begin{itemize}
    \item we filtered the LAION-400M database keeping only image-text pairs where the caption contains a realistic urban scenario;
    \item we extracted all urban objects and the background urban context with the Multimodal LLM (MLLM) Molmo \cite{deitke_molmo_2024};
    \item we generated depth maps using the monocular depth estimation model Depth Anything \cite{depthanything};
    \item we used open vocabulary object detector Grounding DINO \cite{liu2023grounding} to detect objects in the images, yielding bounding boxes;
    \item we developed a metric based on geometric constraints to extract relationships bounding boxes;
    \item we procedurally generate synthetic prompts from the extracted relationships.
\end{itemize}

This pipeline is completely automated and yields a dataset of approximately 5.6 million prompts derived from realistic scenarios of around 22,000 natural images, which we split into training and test sets.
To demonstrate the flexibility of our method, we evaluate it on two open-source state-of-the-art diffusion models with distinct architectures: the UNet-based SDXL and the MMDiT-based Flux.1.
We fine-tune both Flux and SDXL with LoRA on natural data taken from the dataset we built, and synthetic data generated with Flux using our spatially accurate captions.
We then evaluate the models on the test split of the dataset using the same pipeline that we used to build the dataset, extracting relationships from the output images and checking if they correspond to the ones requested by the prompt.
While we focus on urban scenarios as a challenging and practically relevant testbed, our methodology is domain-agnostic: the geometric constraints, synthetic data generation pipeline, and LoRA fine-tuning framework can be readily applied to any domain with spatial relationships.

\begin{figure}[t]
    \centering
    \begin{minipage}[t]{0.46\linewidth}
        \centering
        \includegraphics[width=0.92\linewidth]{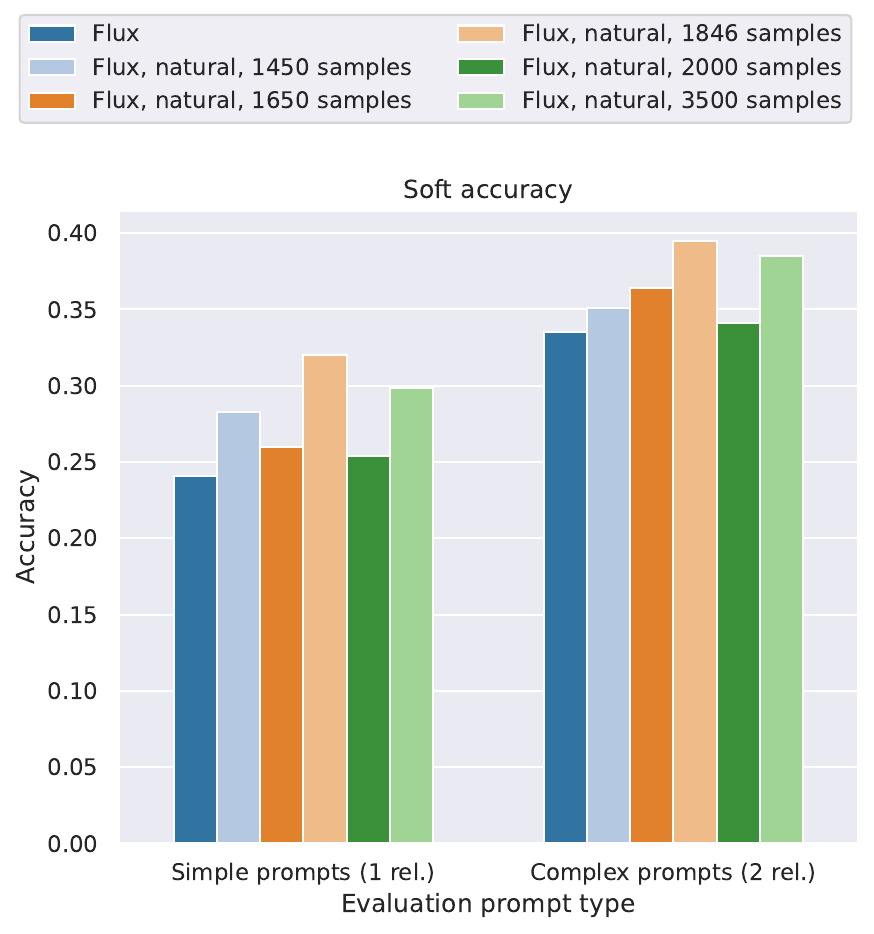}
        \caption{Ablation study on training set size with simple (1 rel.) prompts tested on both simple and complex (2 rel.) prompts containing the \textit{right} relationship. Optimal training set size is around 1800 samples.}
        \label{fig:n-samples}
    \end{minipage}
    \hfill
    \begin{minipage}[t]{0.48\linewidth}
        \centering
        \includegraphics[width=\linewidth]{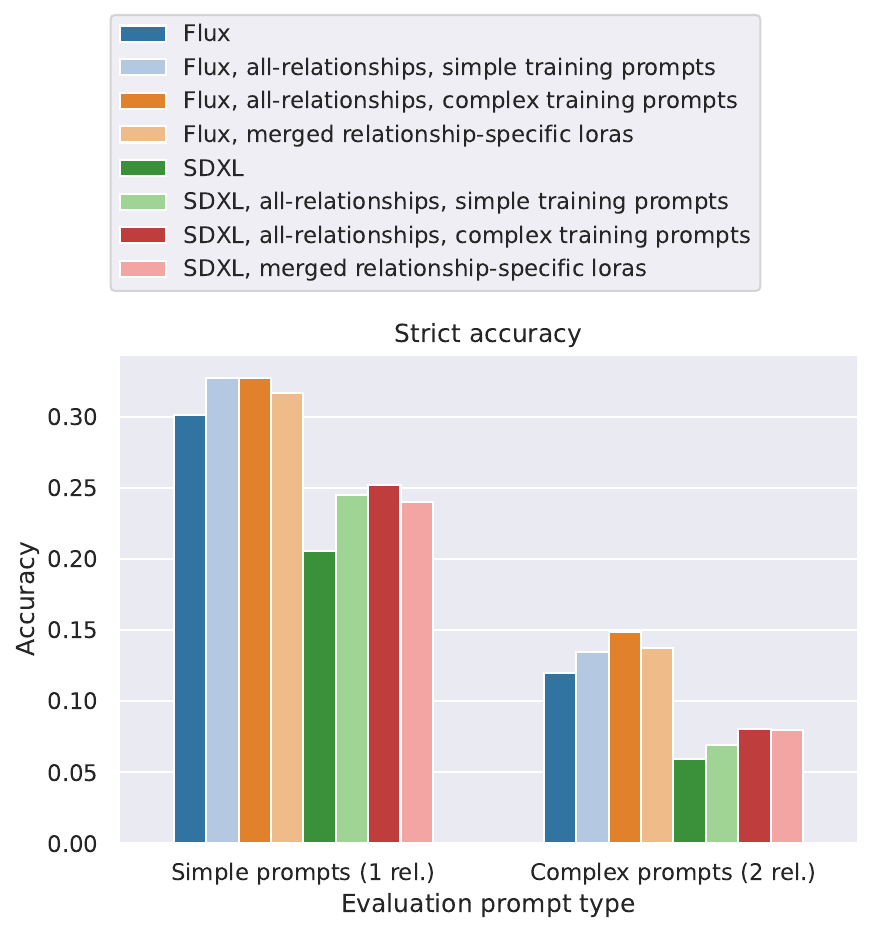}
        \caption{Strict accuracy for all models, trained on synthetic images. Fine-tuning using all relationships and complex (2 rel.) prompts outperforms all other methods.}
        \label{fig:all-strict-acc}
    \end{minipage}
    % \vspace{-0.4cm}
\end{figure}

\begin{table}[t]
\scriptsize
\setlength{\tabcolsep}{6pt}
\centering
\begin{tabular}{l cc @{\hspace{12pt}} | @{\hspace{12pt}} cc @{\hspace{12pt}} | @{\hspace{12pt}} cc}
\toprule
\textbf{} & \textbf{Bottom} & \textbf{Top} & \textbf{Right} & \textbf{Left} & \textbf{Behind} & \textbf{Front} \\ 
\midrule
\textbf{SDXL}            & 0.24 & \textbf{0.25} & 0.16 & 0.16 & 0.24 & \textbf{0.28} \\
\textbf{SDXL + ESPLoRA} & 0.28 & \textbf{0.30} & 0.21 & 0.21 & 0.26 & \textbf{0.30} \\
\midrule
\textbf{Flux.1}          & 0.33 & \textbf{0.41} & 0.31 & \textbf{0.32} & 0.28 & \textbf{0.31} \\
\textbf{Flux.1 + ESPLoRA} & 0.36 & \textbf{0.48} & 0.32 & \textbf{0.34} & 0.30 & \textbf{0.33} \\
\bottomrule
\end{tabular}
\caption{Comparison between opposite relations, averaging accuracies on simple and complex prompts. Most models show a clear bias towards \textit{top, left, front} when compared with the counterparts.}
\label{tab:rel-bias}
\vspace{-0.4cm}
\end{table}

\begin{figure}[t]
    \centering
    \includegraphics[width=0.7\linewidth]{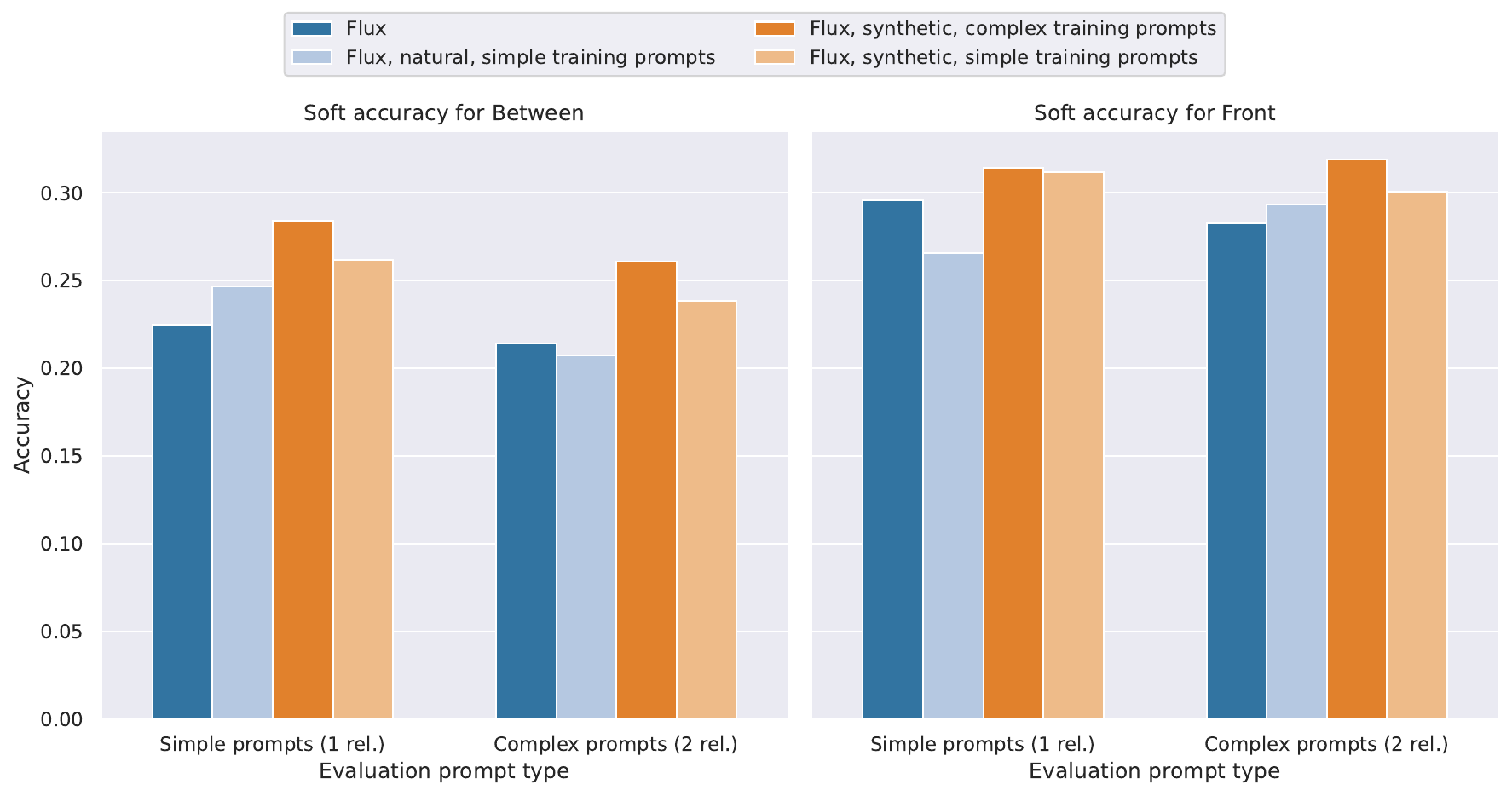}
    \caption{Comparison of soft accuracy on natural and synthetic images for Front and Between. Fine-tuning with synthetic images outperforms all other training configurations.}
    \label{fig:synth-nat-comparison}
    \vspace{-0.4cm}
\end{figure}

\subsection{Extracting relationships from bounding boxes}
\label{sec:rel-metric}
T2I-CompBench \cite{huang_t2i-compbench_2023} introduces a geometric metric for extracting spatial relationships from bounding boxes during evaluation. It consists of a set of geometric constraints on bounding boxes center distances, each corresponding to a specific relationship type.
Given the relationship that the model is expected to generate, the appropriate constraint is applied.
While this might be good enough for evaluation, where the relationships that should appear on the image are known, we find it too ambiguous to extract relationships for training purposes.
In this setting, the relationship present in the image is unknown, requiring the metric to be applied iteratively across all possible relationships. As a result, we often observe that the T2I-CompBench metric assigns objects to multiple relationships simultaneously. We find this ambiguity undesirable in training data, as it could lead to unclear associations between prompts and spatial relationships in the model.
Therefore, we propose an advanced metric based on the box coordinates.

Let \( B_1, B_2, B_3 \) be bounding boxes of three objects, each of which has coordinates: 
$B_i = (x_{\min}^i, y_{\min}^i, x_{\max}^i, y_{\max}^i)$ with $i \in \{1,2,3\}$,
width $\text{w}_i = x_{\max}^i - x_{\min}^i$, and height $\text{h}_i = y_{\max}^i - y_{\min}^i $.
Let \((j,k) = (1,2)\) if \(w_1 \ge w_2\) and \(h_1 \ge h_2\); otherwise,
\((j,k) = (2,1)\).

\vspace{0.4cm}

We define the following distances:
\begin{equation}
    \begin{aligned}
        x_{\max\_dist} &= x_{\max}^j - x_{\max}^k, \quad
        x_{\min\_dist} = x_{\min}^j - x_{\min}^k, \\[4pt]
        y_{\max\_dist} &= y_{\max}^j - y_{\max}^k, \quad
        y_{\min\_dist} = y_{\min}^j - y_{\min}^k.\\
    % \end{aligned}
% \end{equation}
% \begin{equation}
    % \begin{aligned}
        \vspace{0.4cm}\\
        x_{\text{distance}} &= 
        \begin{cases} 
            x_{\min}^1 - x_{\max}^2, & \text{if \textit{locality} = right} \\
            x_{\min}^2 - x_{\max}^1, & \text{if \textit{locality} = left}
        \end{cases} \\
        y_{\text{distance}} &= 
        \begin{cases} 
            y_{\min}^1 - y_{\max}^2, & \text{if \textit{locality} = bottom} \\
            y_{\min}^2 - y_{\max}^1, & \text{if \textit{locality} = top} 
        \end{cases}
    \end{aligned}\\
        \vspace{0.4cm}\\
\end{equation}

We then constrain the relative bounding box positions along the axis orthogonal to the specified \textit{locality}, in order to avoid ambiguities when objects participate in multiple relationships. For \textit{locality} $\in \{\text{right}, \text{left}\}$, we impose the following constraints:
\begin{equation}
\begin{aligned}
    x_{\text{distance}} &> -\frac{\min(w_1, w_2)}{\tau} \\
    y_{\max\_dist} < \frac{\min(h_1, h_2)}{\tau},
    y_{\min\_dist} &> -\frac{\min(h_1, h_2)}{\tau}
    \label{eq:right-pos-constr}
\end{aligned}\\
        \vspace{0.4cm}\\
\end{equation}

The constraints for \textit{locality} $\in \{\text{top}, \text{bottom}\}$ are defined analogously by swapping the $x$ and $y$ axes. The parameter $\tau$ controls the strictness of the constraints.

Taking as example the \textit{right} locality, i.e. "$B_1$ \textit{right} $B_2$", the horizontal constraint in \cref{eq:right-pos-constr} is valid only when the left border of object one ($x_{\min}^1$) is to the right of the right border of object two ($x_{\max}^2$). If that is not the case, the distance between the two, and thus their overlap, must be less than a threshold based on width and a strictness parameter $\tau$.
The vertical constraint instead is only valid when the distance between the upper border of both objects ($y_{\max}^{1}$ and $y_{\max}^{2}$) is less then a threshold based on height and $\tau$.
In all our experiments, we empirically find 3 as the optimal $\tau$.

For the \textit{next} locality we verify that at least one of the \textit{right} and \textit{left} constraints are valid. For the \textit{between} locality instead, we check all possible triplets of bounding boxes, and verify that \textit{left} constraint is valid for $B_1$ and $B_2$ and the \textit{right} constraint is valid for $B_3$ and $B_2$.

For 3D relationships, as shown in \cref{fig:3D-rel-example}, we apply all $x_{\max\_dist},x_{\min\_dist},$ $y_{\max\_dist},y_{\min\_dist}$ constraints to ensure that there is enough overlap between the objects since they should be in front of one another. Then we use Depth Anything to perform monocular depth estimation on the image and we compute average depth $d_1$ for box $B_1$ and $d_2$ for box $B_2$.
If $d_1 > d_2$ then $B_1$ is in front of $B_2$, otherwise the other way around.

\subsection{Transforming Original Relations Effectively (TORE)}
\label{sec:tore}
In our preliminary evaluations, we observed that T2I models perform better in some relationships than in others. This can be seen from our benchmark results in \cref{tab:rel-bias}. The Flux.1 model, as an example, seems to perform better on the \textit{top}, \textit{left}, and \textit{front} relationships than on the corresponding counterparts, \textit{bottom}, \textit{right}, and \textit{behind}.
Such imbalances remain evident even after fine-tuning the models for improved spatial relationship understanding.
Based on this observation, we propose TORE (Transforming Original Relations Effectively), a pre-processing step based on the semantic equivalence of inverse spatial relationships. Recognizing that descriptions like "A right of B" and "B left of A" are semantically identical, TORE detects requested relationships via pattern matching. If a prompt targets a relationship where the model is known to be weak, TORE flips it to the corresponding higher-performing variant (e.g., transforming "A bottom B" into "B top A"). This approach increases generation accuracy without changing the prompt meaning, enhancing performance whether used with base T2I models or alongside ESPLoRA.

\begin{table}[t]
\scriptsize
\setlength{\tabcolsep}{2pt}
\centering
\begin{tabular}{lcc @{\hspace{15pt}} | @{\hspace{15pt}} lcc}
\toprule
\textbf{} & \begin{tabular}[c]{@{}c@{}}\textbf{Textual}\\\textbf{Variations}\end{tabular} & \begin{tabular}[c]{@{}c@{}}\textbf{T. Non}\\\textbf{Spatial}\end{tabular} & \textbf{} & \begin{tabular}[c]{@{}c@{}}\textbf{Textual}\\\textbf{Variations}\end{tabular} & \begin{tabular}[c]{@{}c@{}}\textbf{T. Non}\\\textbf{Spatial}\end{tabular} \\ 
\midrule
\textbf{SDXL} & 0.15 & 0.31 & \textbf{Flux.1} & 0.23 & 0.30 \\
\textbf{SDXL + ESPLoRA} & 0.17 & 0.31 & \textbf{Flux.1 + ESPLoRA} & 0.26 & 0.30 \\
\bottomrule
\end{tabular}
\caption{Results on textual variations and non-spatial prompts. Our approach is robust to alternative textual representations of relationships, and retains prompt following capabilities on non-spatial scenarios.}
\label{tab:test}
\vspace{-0.7cm}
\end{table}

\section{The Urban T2I Spatial Dataset}
\label{sec:dataset}
To create the proposed Urban T2I Spatial Dataset, we extracted a list of 299 urban objects from real-world urban design documents, we extend it with Molmo, and filter LAION to captions mentioning at least one object.
Instead of simply picking a set of objects and generating every possible relationship between them, we start with real-world documents to extract genuine spatial relationships. This ensures our data reflects natural human environments and logical object placements, whereas objects permutations often result in layouts that don't exist in reality.
Using Molmo, Grounding DINO, and Depth Anything, we derive 2D and 3D spatial relations between detected boxes as described in \cref{sec:rel-metric}, obtaining \texttt{<object, relationship, object, context>} quadruples.

We focus on four urban contexts (\emph{city}, \emph{street}, \emph{downtown area}, \emph{residential area}) and augment relations by inversion (e.g. \texttt{<A, right, B, context>} can be converted to \texttt{<B, left, A, context>} and retain the same meaning.). We procedurally generate simple (one relation) and complex (two relations) prompts, producing 5.6M prompts over 22K LAION images, with a separate evaluation split consisting of 14,720 prompts (balanced by relation type and context). Inspired by \cite{zhang_compass_2024}, we impose object-proximity and cropping constraints to strengthen spatial signals. 
Moreover, we create a second dataset by generating images with Flux.1 from our prompts and re-validating them with the same pipeline: only images satisfying all prompted relations are kept. This synthetic set is smaller due to low relation satisfaction rates of the Flux.1 base model. This has the additional benefit of avoiding low-quality images.
The double validation allows us to mitigate potential inaccuracies introduced by object detection models and spatial constraints. Each step in the pipeline progressively refines the data: natural-image filtering reduces noisy captions, while synthetic-image verification ensures that only relations faithfully satisfied in the visual domain are retained. This layered validation strategy  decreases the risk of error and enhances the precision of the resulting dataset. In this way, the two released datasets (natural and synthetic) complement each other while providing reliable and high-quality urban spatial relations.

\section{Experimental Setting}
\label{sec:exp-set}
In this work, we explore ways to improve the spatial understanding capabilities of Flux.1 and SDXL models, chosen for their ability to produce high-quality outputs at resolutions of 1024$\times$1024, consistent with modern generative standards. To achieve this, we train a variety of LoRA models using different prompts and LoRA configurations.

For prompts, we explore two different prompt configurations. Prompts with only one relationship (1 rel.), which we name simple prompts, and the more challenging prompts with two relationships (2 rel.), named complex prompts.
In an attempt to further improve performance, we explore merging the LoRA models trained on different relationships.
We select the best-performing model for each individual relationship based on validation performance and combine them into a unified LoRA model. Finally, we evaluate all the trained models on the test split of our dataset.
We also apply the TORE preprocessing step to investigate its effectiveness.

\subsection{Evaluation Metrics}
\label{sec:eval-metrics}
For evaluation, we consider two different evaluation metrics:
\begin{itemize}
    \item To evaluate text-image alignment capabilities, we use \textbf{T2ICompBench} \cite{huang_t2i-compbench_2023}. It covers attribute binding, counting and spatial and non-spatial relationships across diverse non-urban scenarios. This confirms that the spatial understanding learned from our urban-focused training method transfers successfully to broader, general domains.
    
    \item A novel proposed \textbf{Urban Benchmark}: We generate images for the test split of our urban-focused dataset, which is strictly disjoint from the training data, resulting in a total of 14,720 evaluation text-image pairs. We apply the same pipeline and geometrical constraints explained in \cref{sec:methodology} to the images generated using the models. 
    We assign a score of 1 to each relationship in the input prompt if the corresponding bounding boxes satisfy the geometrical constraints for the target relationship.
    We define two metrics: soft accuracy and strict accuracy. For \textit{soft accuracy} we select a specific relationship we want to evaluate and count the sample as correct (score 1) if that relationship was correctly rendered in the image, ignoring the score of any other relationship that might be present in the prompt.
    For \textit{strict accuracy} we count the sample correct if all relationships in the prompt are found correctly rendered by the constraints.
\end{itemize}

By coupling our in-domain, geometry-aware Urban Benchmark with the well-established, general-domain T2ICompBench, we rigorously evaluate our approach, showing improved urban spatial grounding and generalization without sacrificing overall text–image alignment.

\begin{table}[t]
\scriptsize
\setlength{\tabcolsep}{5pt}
\centering
\begin{tabular}{lccc}
\toprule
\textbf{Model} & \begin{tabular}[c]{@{}c@{}} \textbf{Image} \textbf{Size}\end{tabular} & \begin{tabular}[c]{@{}c@{}}\textbf{T. Spatial} \textbf{2D}\end{tabular} & \begin{tabular}[c]{@{}c@{}}\textbf{T. Spatial} \textbf{3D}\end{tabular}\\ \midrule
SDXL & 1024\textsuperscript{2} & 0.20 & 0.36 \\
SDXL + TORE & 1024\textsuperscript{2} & 0.21 & 0.37 \\
SDXL + ESPLoRA (1 rel.) & 1024\textsuperscript{2} & 0.25 & 0.37 \\
SDXL + ESPLoRA (2 rel.) & 1024\textsuperscript{2} & 0.23 & 0.37 \\
SDXL + ESPLoRA (1 rel.) + TORE & 1024\textsuperscript{2} & 0.25 & 0.38 \\
SDXL + ESPLoRA (2 rel.) + TORE & 1024\textsuperscript{2} & 0.24 & 0.38 \\
\midrule
Flux.1 & 512\textsuperscript{2} & 0.18 & - \\
Flux.1 & 1024\textsuperscript{2} & 0.26 & 0.36 \\
Flux.1 + TORE & 1024\textsuperscript{2} & 0.28 & 0.39 \\
Flux.1 + CoMPaSS & 512\textsuperscript{2} & 0.30 & - \\
Flux.1 + ESPLoRA (1 rel.) & 1024\textsuperscript{2} & 0.31 & 0.39 \\
Flux.1 + ESPLoRA (2 rel.) & 1024\textsuperscript{2} & 0.30 & 0.40 \\
Flux.1 + ESPLoRA (1 rel.) + TORE & 1024\textsuperscript{2} & 0.31 & 0.40 \\
\textbf{Flux.1 + ESPLoRA (2 rel.) + TORE} & 1024\textsuperscript{2} & \textbf{0.34} & \textbf{0.42}\\
\bottomrule
\end{tabular}
\caption{Results on T2ICompBench. Our best model outperforms both the base model and CoMPaSS \cite{zhang_compass_2024}, setting a new state-of-the-art on the evaluated spatial benchmark.}
\label{tab:bech-results}
    \vspace{-0.5cm}
\end{table}

\begin{figure}[t]
    \centering
    \includegraphics[width=1.0\linewidth]{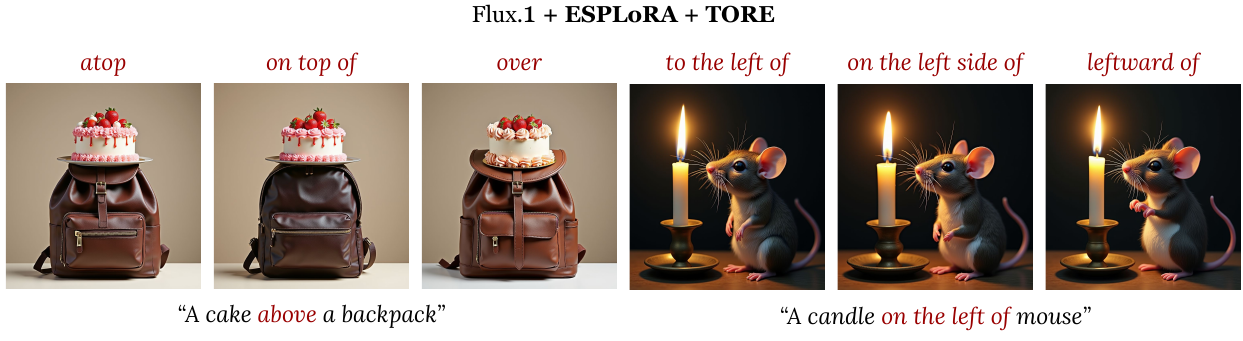}
    \caption{Qualitative evaluation on textual variations to show  robustness to alternative textual representations of relationships.}
    \label{fig:text-variations}
    \vspace{-0.4cm}
\end{figure}

\begin{figure}[t]
    \centering
    \begin{minipage}[t]{0.31\linewidth}
        \centering
        \includegraphics[width=0.97\linewidth]{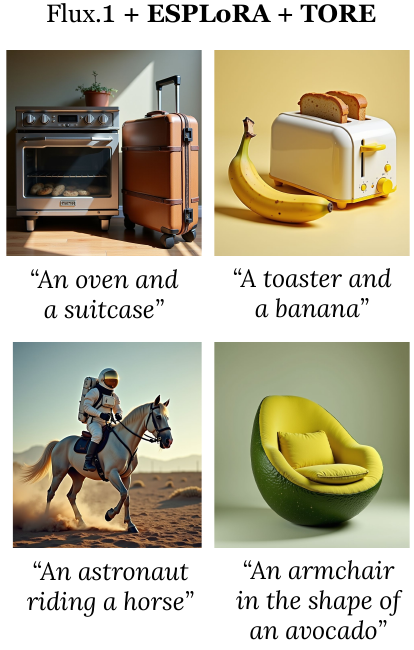}
        \caption{Qualitative evaluation on non-spatial prompts demonstrating robustness with general-purpose prompt.}
        \label{fig:non-spatial}
    \end{minipage}
    \hfill
    \begin{minipage}[t]{0.63\linewidth}
        \centering
        \includegraphics[width=1.0\linewidth]{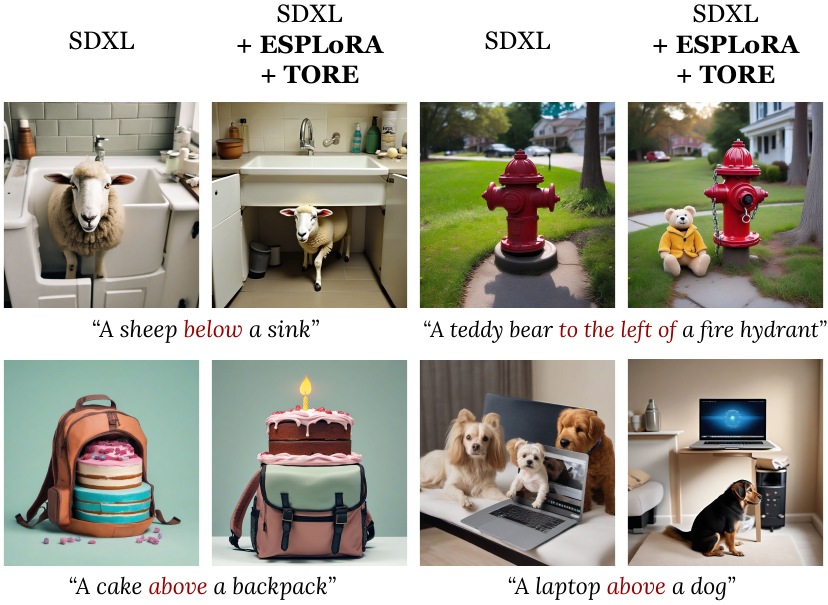}
        \caption{Qualitative comparison on traditional one-relationship spatial prompts. SDXL + ESPLoRA + TORE outperforms SDXL base model generating accurate spatial layouts. }
        \label{fig:sdxl-vs-us}
    \end{minipage}
    \vspace{-0.4cm}
\end{figure}

\section{Results}
\label{sec:results}

\begin{figure}[t]
    \centering
    \includegraphics[width=1\linewidth]{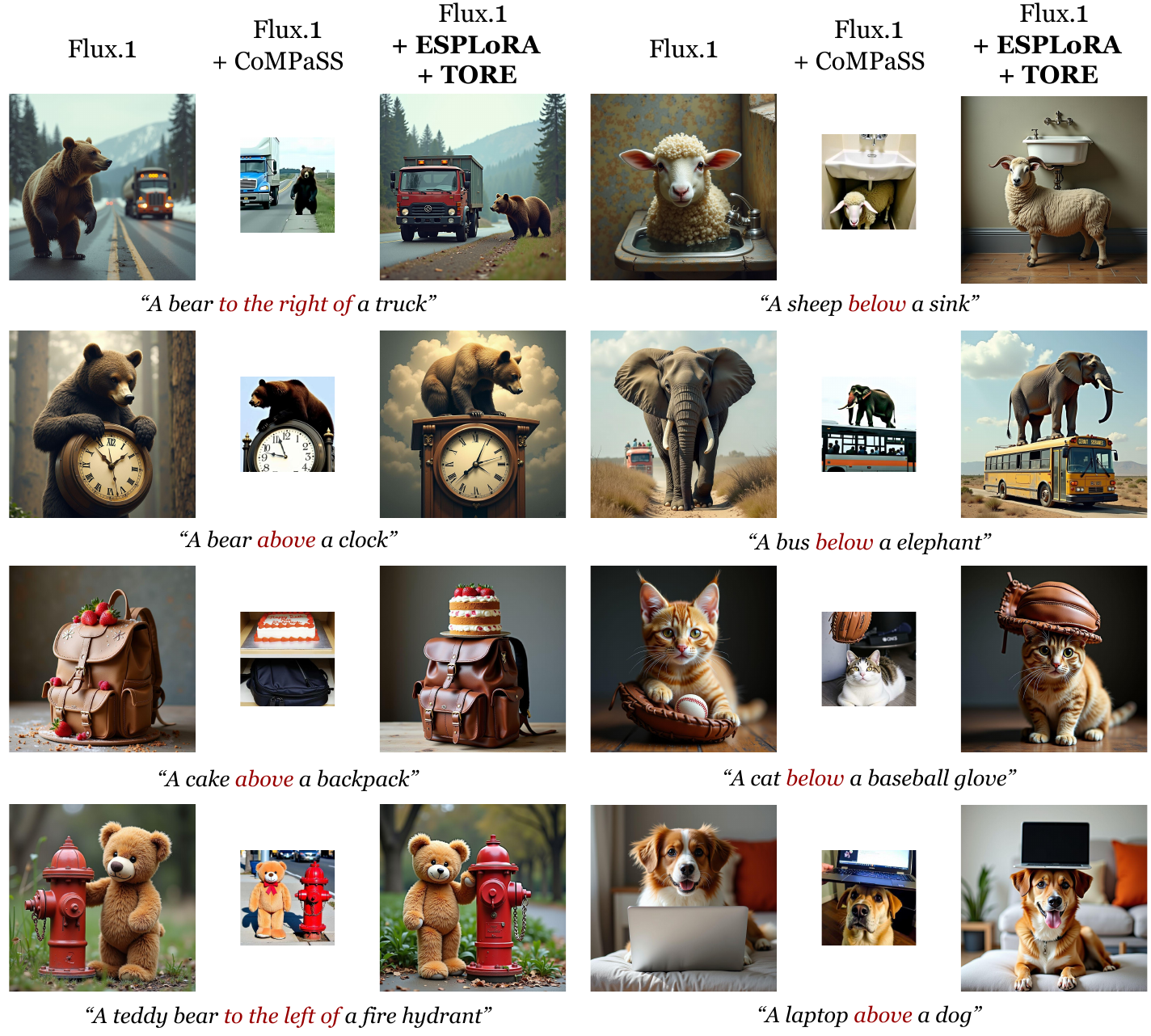}
    \caption{Qualitative comparison on traditional one-relationship spatial prompts. CoMPaSS improves spatial accuracy, but it generates low-res, cropped images. ESPLoRA + TORE achieves both accurate spatial layouts and hd image quality. }
    \label{fig:compass-vs-us}
    \vspace{-0.4cm}
\end{figure}

\begin{figure}[t]
    \centering
    \includegraphics[width=1\linewidth]{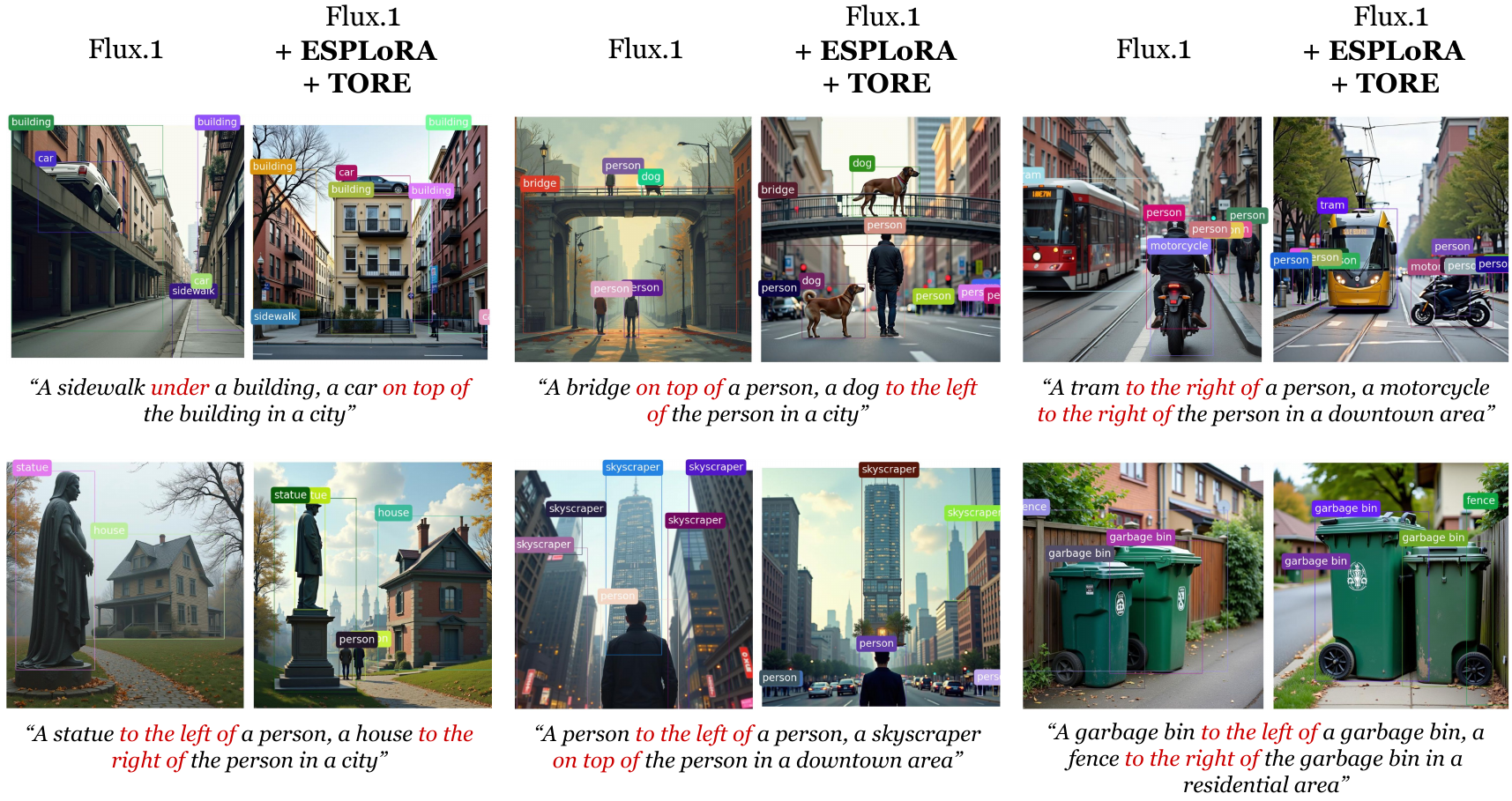}
    \caption{Qualitative comparison between Flux.1 and Flux.1 + ESPLoRA + TORE on spatial prompts from our Urban Benchmark.}
    \label{fig:2r_qual}
    \vspace{-0.4cm}
\end{figure}

First we train relationship-specific LoRAs on the cropped natural images from our dataset, with simple (one relationship) prompts.
Although we do train models on complex (two relationship) prompts and natural images, we find their performance to always be on par or worse than the base model.
We ablate the training-set size by randomly sampling 3500, 2000, 1846, 1650, and 1450 images, training LoRA models as in \cref{sec:exp-set}, and evaluating on our Urban Benchmark. As shown in \cref{fig:n-samples}, soft accuracy for the \textit{right} relation peaks around 1.8K samples (1846), with diminishing returns beyond that point. 
We then compare natural versus synthetic training data using relationship-specific LoRAs: \cref{fig:synth-nat-comparison} shows that fine-tuning on \emph{synthetic} images achieves the highest soft accuracy for \textit{front} and \textit{between}, outperforming all natural-image configurations and the base model. Based on this clear advantage, we conduct all subsequent experiments with synthetic data.

We train relationship-specific models for each relationship and all-relationship models with synthetic data on all relationships, on both prompt varieties. We also select our best relationship-specific models and merge them.
We show the result of all the cited models on our benchmark in \cref{fig:all-strict-acc} and \cref{tab:rel-bias}.
As shown in \cref{fig:all-strict-acc}, LoRAs trained on all relationships always outperform the base model, and models trained on complex (2 rel.) prompts on all relationships outperform merged relationship-specific models.
\cref{tab:rel-bias} shows that models are biased and perform better on \textit{top, left, front} than on the counterparts \textit{bottom, right, behind}, which lead us to develop TORE, as described in \cref{sec:tore}. Furthermore, as shown in \cref{tab:test}, \cref{fig:text-variations}, and \cref{fig:non-spatial}, ESPLoRA is robust to textual paraphrases of relations while keeping non-spatial prompt following unchanged, indicating that our fine-tuning improves relational understanding without overfitting to specific phrasings or degrading general capabilities.
\cref{fig:sdxl-vs-us}, \cref{fig:compass-vs-us}, and \cref{fig:2r_qual} show a qualitative comparison of the base SDXL model, the base Flux.1 model, CoMPaSS method, and our method. Our method helps the model accurately render spatial relationships while retaining high-definition quality.

We evaluate our best performers, Flux.1 and SDXL with all-relationship LoRAs trained on simple (1 rel.) and complex (2 rel.) prompts respectively, on T2ICompBench 2D and 3D and report results in \cref{tab:bech-results}, compared with CoMPaSS \cite{zhang_compass_2024}.
We also evaluate our overall best models on the Non spatial split of T2ICompBench, showing that our fine-tuning framework can improve spatial accuracy while preserving the model's general prompt following capabilities.
Additionally, we investigate wether our fine-tuning strategy may bias the model towards the specific spatial prompts used during training.
We reformulate T2ICompBench 2D prompts using three different textual variations for each relationship and use the new prompts for evaluation. For example, we replace "on the bottom of" with "below", "underneath" or "under". Since the models are not trained on the new textual forms the performance drops as expected. Nevertheless, our models outperform their respective base models, indicating improved generalization.
The consistent improvements observed on both the urban benchmark and the independent T2I-CompBench indicate genuine spatial understanding rather than overfitting to specific data distributions.

Finally we apply our novel pre-processing step TORE to our best models.
Our training method, based on synthetic images and spatially accurate prompts, outperforms both Flux.1 and the state-of-the-art method, CoMPaSS, on T2ICompBench 2D and 3D.

\section{Conclusion}
\label{sec:conclusion}
In this work, we introduced ESPLoRA, a novel fine-tuning framework based on Low-Rank Adaptation, designed to enhance the spatial consistency of text-to-image diffusion models. By harnessing a curated dataset with spatially explicit prompts, ESPLoRA significantly improves the alignment between textual descriptions and visual outputs, outperforming current state-of-the-art methods while maintaining high-definition outputs quality and minimal inference time overhead. Furthermore, we proposed a novel, geometry-grounded evaluation metric and benchmark, which provide rigorous means to assess and quantify spatial coherence in generated images, including 3D spatial relations in images such as \textit{in front of} or \textit{behind}. Our fine-tuning framework together with the TORE algorithm, which exploits existing model biases, lead to a 13.33\% performance improvement over previous approaches. The results demonstrate that our method significantly enhances the ability of T2I models to render complex spatial constraints involving multiple relationships between objects, offering promising benefits for applications such as urban planning and design. Importantly, these improvements come without any significant increase in inference time or loss in output quality.

\bibliographystyle{splncs04}
\bibliography{main,bibzt}

@misc{brooks_instructpix2pix_2023,
	title = {{InstructPix2Pix}: {Learning} to {Follow} {Image} {Editing} {Instructions}},
	shorttitle = {{InstructPix2Pix}},
	publisher = {arXiv},
	author = {Brooks, Tim and Holynski, Aleksander and Efros, Alexei A.},
	year = {2023},
	note = {arXiv:2211.09800}
}

@misc{huang_t2i-compbench_2023,
	title = {{T2I}-{CompBench}: {A} {Comprehensive} {Benchmark} for {Open}-world {Compositional} {Text}-to-image {Generation}},
	shorttitle = {{T2I}-{CompBench}},
	publisher = {arXiv},
	author = {Huang, Kaiyi and Sun, Kaiyue and et al.},
	year = {2023},
	note = {arXiv:2307.06350}
}

@misc{deitke_molmo_2024,
	title = {Molmo and {PixMo}: {Open} {Weights} and {Open} {Data} for {State}-of-the-{Art} {Vision}-{Language} {Models}},
	shorttitle = {Molmo and {PixMo}},
	publisher = {arXiv},
	author = {Deitke, Matt and Clark, Christopher and Lee, Sangho and et al.},
	year = {2024},
	note = {arXiv:2409.17146}
}

@misc{phung_grounded_2023,
	title = {Grounded {Text}-to-{Image} {Synthesis} with {Attention} {Refocusing}},
	publisher = {arXiv},
	author = {Phung, Quynh and Ge, Songwei and Huang, Jia-Bin},
	year = {2023},
	note = {arXiv:2306.05427}
}

@misc{gokhale_benchmarking_2023,
	title = {Benchmarking {Spatial} {Relationships} in {Text}-to-{Image} {Generation}},
	publisher = {arXiv},
	author = {Gokhale, Tejas and Palangi, Hamid and Nushi, Besmira and Vineet, Vibhav and Horvitz, Eric and et al.},
	year = {2023},
	note = {arXiv:2212.10015}
}

@misc{ghosh_geneval_2023,
	title = {{GenEval}: {An} {Object}-{Focused} {Framework} for {Evaluating} {Text}-to-{Image} {Alignment}},
	shorttitle = {{GenEval}},
	publisher = {arXiv},
	author = {Ghosh, Dhruba and Hajishirzi, Hanna and Schmidt, Ludwig},
	year = {2023},
	note = {arXiv:2310.11513}
}

@misc{chatterjee_getting_2024,
	title = {Getting it {Right}: {Improving} {Spatial} {Consistency} in {Text}-to-{Image} {Models}},
	shorttitle = {Getting it {Right}},
	publisher = {arXiv},
	author = {Chatterjee, Agneet and Stan, Gabriela Ben Melech and Aflalo, Estelle and Paul, Sayak and et al.},
	year = {2024},
	note = {arXiv:2404.01197},
}

@misc{zhang_compass_2024,
	title = {{CoMPaSS}: {Enhancing} {Spatial} {Understanding} in {Text}-to-{Image} {Diffusion} {Models}},
	shorttitle = {{CoMPaSS}},
	publisher = {arXiv},
	author = {Zhang, Gaoyang and Fu, Bingtao and Fan, Qingnan and Zhang, Qi and Liu, Runxing and Gu, Hong and et al.},
	year = {2024},
	note = {arXiv:2412.13195}
}

@misc{hu_ella_2024,
	title = {{ELLA}: {Equip} {Diffusion} {Models} with {LLM} for {Enhanced} {Semantic} {Alignment}},
	shorttitle = {{ELLA}},
	publisher = {arXiv},
	author = {Hu, Xiwei and Wang, Rui and Fang, Yixiao and Fu, Bin and Cheng, Pei and Yu, Gang},
	year = {2024},
	note = {arXiv:2403.05135}
}

@misc{wu_paragraph--image_2023,
	title = {Paragraph-to-{Image} {Generation} with {Information}-{Enriched} {Diffusion} {Model}},
	publisher = {arXiv},
	author = {Wu, Weijia and Li, Zhuang and He, Yefei and Shou, Mike Zheng and Shen, Chunhua and et al.},
	year = {2023},
	note = {arXiv:2311.14284},
}

@misc{koh_generating_2023,
	title = {Generating {Images} with {Multimodal} {Language} {Models}},
	publisher = {arXiv},
	author = {Koh, Jing Yu and Fried, Daniel and Salakhutdinov, Ruslan},
	year = {2023},
	note = {arXiv:2305.17216},
}

@misc{fu_guiding_2024,
	title = {Guiding {Instruction}-based {Image} {Editing} via {Multimodal} {Large} {Language} {Models}},
	publisher = {arXiv},
	author = {Fu, Tsu-Jui and Hu, Wenze and Du, Xianzhi and Wang, William Yang and et al.},
	year = {2024},
	note = {arXiv:2309.17102}
}

@misc{zhao_uni-controlnet_2023,
	title = {Uni-{ControlNet}: {All}-in-{One} {Control} to {Text}-to-{Image} {Diffusion} {Models}},
	shorttitle = {Uni-{ControlNet}},
	publisher = {arXiv},
	author = {Zhao, Shihao and Chen, Dongdong and Chen, Yen-Chun and Bao, Jianmin and Hao, Shaozhe and et al.},
	year = {2023},
	note = {arXiv:2305.16322}
}

@misc{liu_visual_2023,
	title = {Visual {Instruction} {Tuning}},
	publisher = {arXiv},
	author = {Liu, Haotian and Li, Chunyuan and Wu, Qingyang and Lee, Yong Jae},
	year = {2023},
	note = {arXiv:2304.08485},
}

@misc{mou_t2i-adapter_2023,
	title = {{T2I}-{Adapter}: {Learning} {Adapters} to {Dig} out {More} {Controllable} {Ability} for {Text}-to-{Image} {Diffusion} {Models}},
	shorttitle = {{T2I}-{Adapter}},
	publisher = {arXiv},
	author = {Mou, Chong and Wang, Xintao and Xie, Liangbin and et al.},
	year = {2023},
	note = {arXiv:2302.08453}
}

@misc{zhang_controllable_2023,
	title = {Controllable {Text}-to-{Image} {Generation} with {GPT}-4},
	publisher = {arXiv},
	author = {Zhang, Tianjun and Zhang, Yi and Vineet, Vibhav and Joshi, Neel and Wang, Xin},
	year = {2023},
	note = {arXiv:2305.18583}
}

@misc{chefer_attend-and-excite_2023,
	title = {Attend-and-{Excite}: {Attention}-{Based} {Semantic} {Guidance} for {Text}-to-{Image} {Diffusion} {Models}},
	shorttitle = {Attend-and-{Excite}},
	publisher = {arXiv},
	author = {Chefer, Hila and Alaluf, Yuval and Vinker, Yael and et al.},
	year = {2023},
	note = {arXiv:2301.13826}
}

@misc{lian_llm-grounded_2023,
	title = {{LLM}-grounded {Diffusion}: {Enhancing} {Prompt} {Understanding} of {Text}-to-{Image} {Diffusion} {Models} with {Large} {Language} {Models}},
	shorttitle = {{LLM}-grounded {Diffusion}},
	publisher = {arXiv},
	author = {Lian, Long and Li, Boyi and et al.},
	year = {2023},
	note = {arXiv:2305.13655}
}

@misc{li_gligen_2023,
	title = {{GLIGEN}: {Open}-{Set} {Grounded} {Text}-to-{Image} {Generation}},
	shorttitle = {{GLIGEN}},
	publisher = {arXiv},
	author = {Li, Yuheng and Liu, Haotian and Wu, Qingyang and Mu, Fangzhou and Yang, Jianwei and Gao, Jianfeng and Li, Chunyuan and Lee, Yong Jae},
	year = {2023},
	note = {arXiv:2301.07093},
}

@misc{feng_layoutgpt_2023,
	title = {{LayoutGPT}: {Compositional} {Visual} {Planning} and {Generation} with {Large} {Language} {Models}},
	shorttitle = {{LayoutGPT}},
	publisher = {arXiv},
	author = {Feng, Weixi and Zhu, Wanrong and Fu, Tsu-jui and Jampani, Varun and et al.},
	year = {2023},
	note = {arXiv:2305.15393}
}

@misc{wu_self-correcting_2023,
	title = {Self-correcting {LLM}-controlled {Diffusion} {Models}},
	publisher = {arXiv},
	author = {Wu, Tsung-Han and Lian, Long and Gonzalez, Joseph E. and Li, Boyi and Darrell, Trevor},
	year = {2023},
	note = {arXiv:2311.16090}
}

@misc{zhang_adding_2023,
	title = {Adding {Conditional} {Control} to {Text}-to-{Image} {Diffusion} {Models}},
	publisher = {arXiv},
	author = {Zhang, Lvmin and Rao, Anyi and Agrawala, Maneesh},
	year = {2023},
	note = {arXiv:2302.05543}
}

@misc{dhariwal_diffusion_2021,
	title = {Diffusion {Models} {Beat} {GANs} on {Image} {Synthesis}},
	publisher = {arXiv},
	author = {Dhariwal, Prafulla and Nichol, Alex},
	year = {2021},
	note = {arXiv:2105.05233},
}

@misc{yang2022reco,
      title={ReCo: Region-Controlled Text-to-Image Generation}, 
      author={Zhengyuan Yang and Jianfeng Wang and Zhe Gan and Linjie Li and Kevin Lin and Chenfei Wu and Nan Duan and Zicheng Liu and Ce Liu and Michael Zeng and Lijuan Wang},
      year={2022},
      eprint={2211.15518},
      archivePrefix={arXiv},
      primaryClass={cs.CV},
}

@misc{zhao2024bridging,
      title={Bridging Different Language Models and Generative Vision Models for Text-to-Image Generation}, 
      author={Shihao Zhao and Shaozhe Hao and Bojia Zi and Huaizhe Xu and Kwan-Yee K. Wong},
      year={2024},
      note={arXiv 2403.07860}
}

@misc{qin2024diffusiongpt,
      title={DiffusionGPT: LLM-Driven Text-to-Image Generation System}, 
      author={Jie Qin and Jie Wu and Weifeng Chen and Yuxi Ren and Huixia Li and Hefeng Wu and Xuefeng Xiao and Rui Wang and Shilei Wen},
      year={2024},
      eprint={2401.10061},
      archivePrefix={arXiv},
      primaryClass={cs.CV}
}

@misc{schuhmann2021laion400mopendatasetclipfiltered,
      title={LAION-400M: Open Dataset of CLIP-Filtered 400 Million Image-Text Pairs}, 
      author={Christoph Schuhmann and Richard Vencu and Romain Beaumont and Robert Kaczmarczyk and et al.},
      year={2021},
      note={arXiv 2111.02114},
}

@misc{lin2015microsoftcococommonobjects,
      title={Microsoft COCO: Common Objects in Context}, 
      author={Tsung-Yi Lin and Michael Maire and Serge Belongie and Lubomir Bourdev and Ross Girshick and James Hays and Pietro Perona and et al.},
      year={2015},
      note={arXiv 1405.0312},
}

@article{liu2023grounding,
  title={Grounding dino: Marrying dino with grounded pre-training for open-set object detection},
  author={Liu, Shilong and Zeng, Zhaoyang and et al.},
  journal={arXiv preprint arXiv:2303.05499},
  year={2023}
}

@inproceedings{depthanything,
      title={Depth Anything: Unleashing the Power of Large-Scale Unlabeled Data}, 
      author={Yang, Lihe and Kang, Bingyi and Huang, Zilong and Xu, Xiaogang and Feng, Jiashi and Zhao, Hengshuang},
      booktitle={CVPR},
      year={2024}
}

@inproceedings{rombach2022high,
  title={High-resolution image synthesis with latent diffusion models},
  author={Rombach, Robin and Blattmann, Andreas and Lorenz, Dominik and Esser, Patrick and Ommer, Bj{\"o}rn},
  booktitle={CVPR},
  year={2022}
}

@misc{flux2024,
    author={Black Forest Labs},
    title={FLUX},
    year={2024},
    howpublished={\url{https://github.com/black-forest-labs/flux}},
}

@article{hu2022lora,
  title={Lora: Low-rank adaptation of large language models.},
  author={Hu, Edward J and Shen, Yelong and Wallis, Phillip and Allen-Zhu, Zeyuan and Li, Yuanzhi and Wang, Shean and Wang, Lu and Chen, Weizhu and others},
  journal={ICLR},
  volume={1},
  number={2},
  pages={3},
  year={2022}
}

% \clearpage
\appendix
\section{Qualitatives}
\begin{figure}[t]
    \centering
    \includegraphics[width=0.7\linewidth]{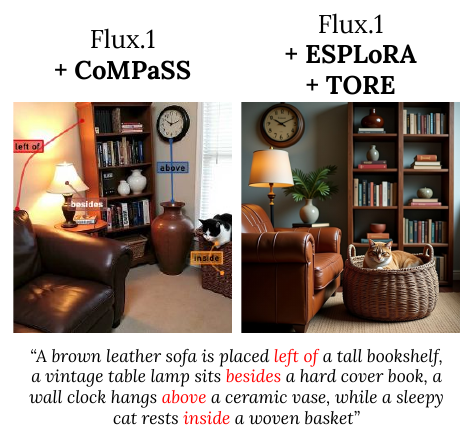}
    \caption{Comparison with CoMPaSS on an indoor scenario with four relationships.}
    \label{fig:4-rel-comparison}
\end{figure}

\begin{figure}[h]
    \centering
    \includegraphics[width=0.6\linewidth]{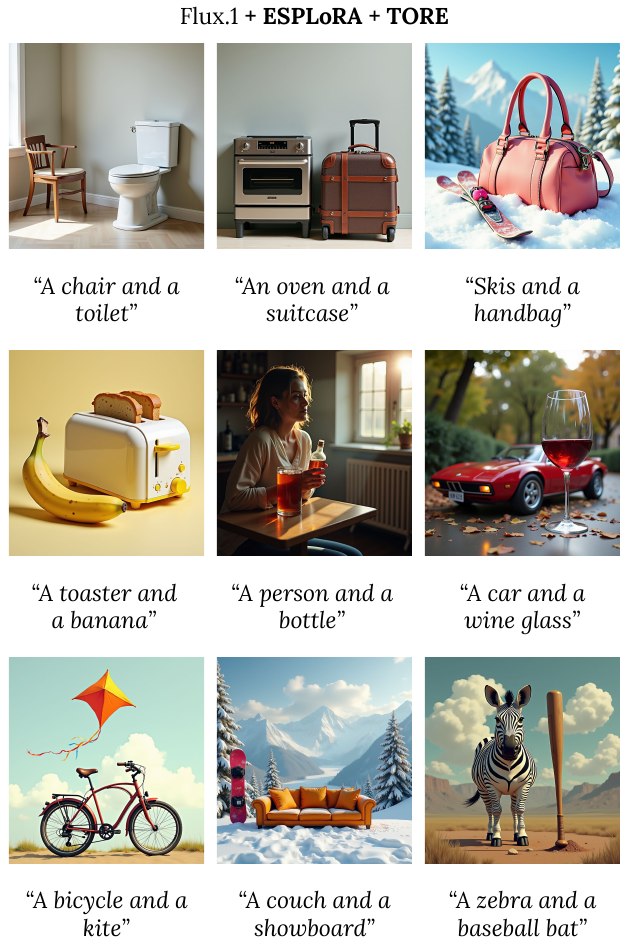}
    \caption{Qualitatives for Non spatial T2ICompBench}
    \label{fig:non-spatial-quals}
\end{figure}

\begin{figure*}[h]
    \centering
    \includegraphics[width=\linewidth]{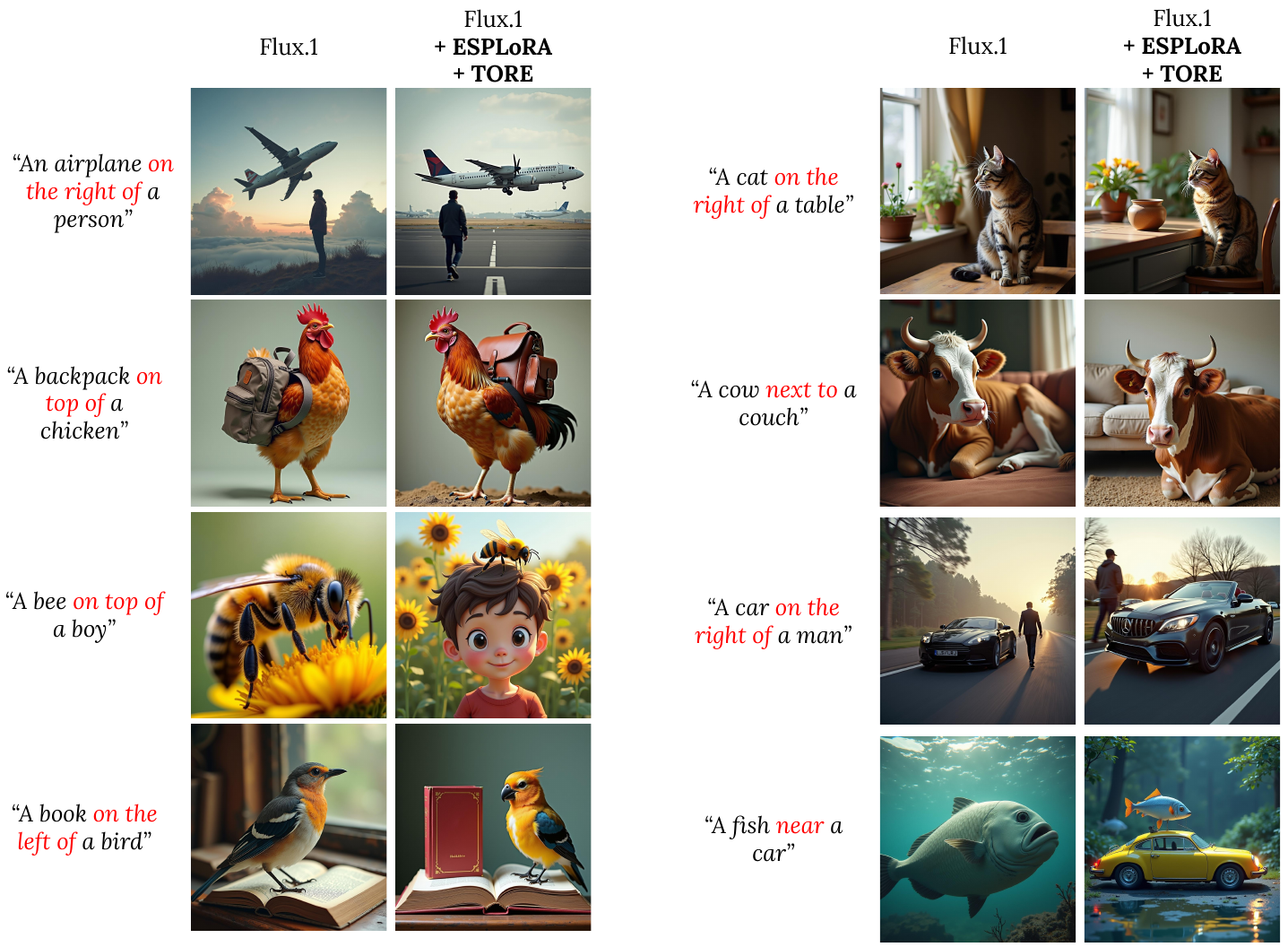}
    \caption{Additional qualitatives for our method.}
    \label{fig:extra-quals-flux}
\end{figure*}

\begin{figure*}[h]
    \centering
    \includegraphics[width=\linewidth]{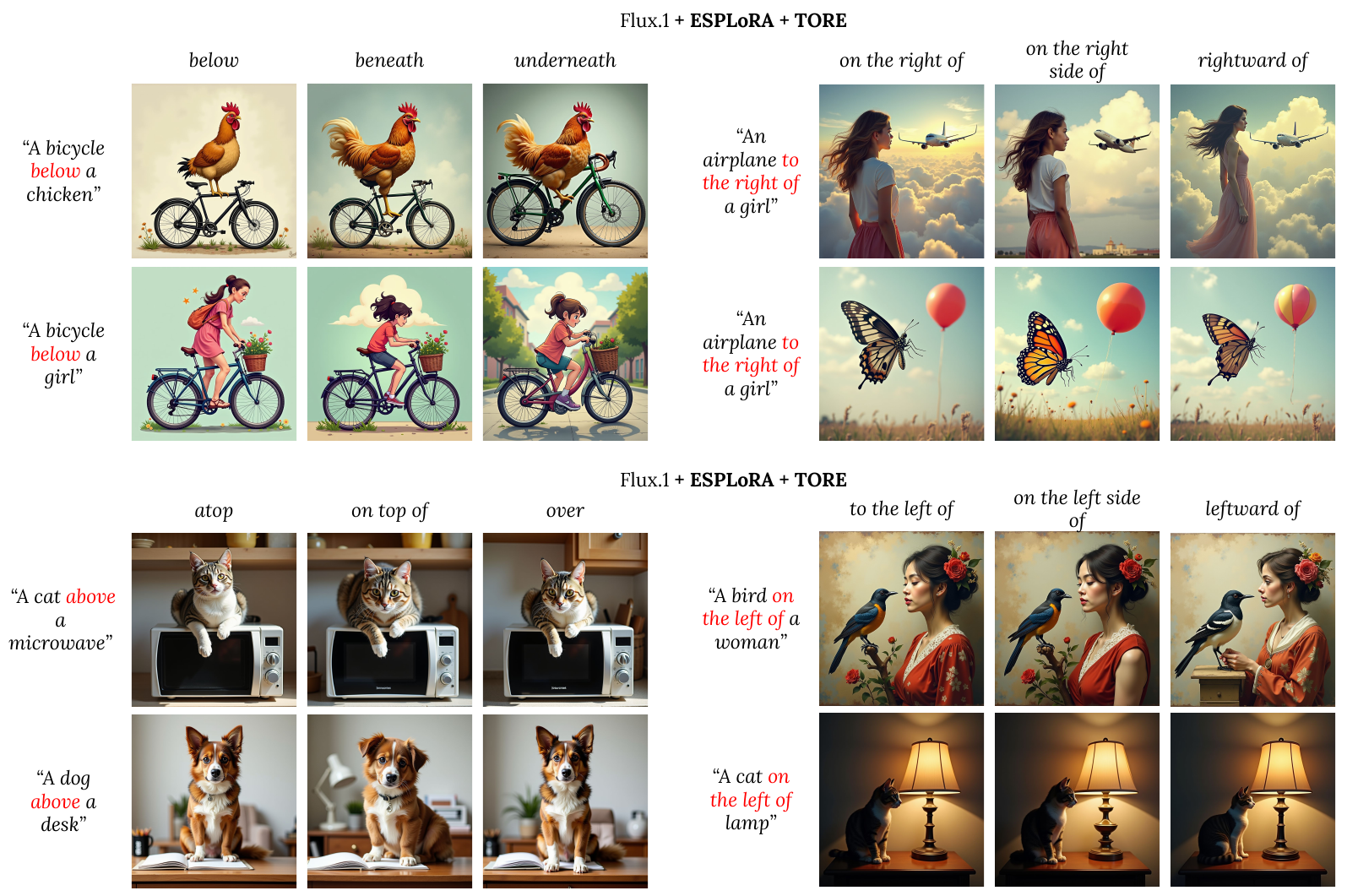}
    \caption{Qualitatives for textual variations on T2ICompBench 2D}
    \label{fig:variations-quals}
\end{figure*}

\begin{figure*}[t]
    \centering
    \includegraphics[width=\linewidth]{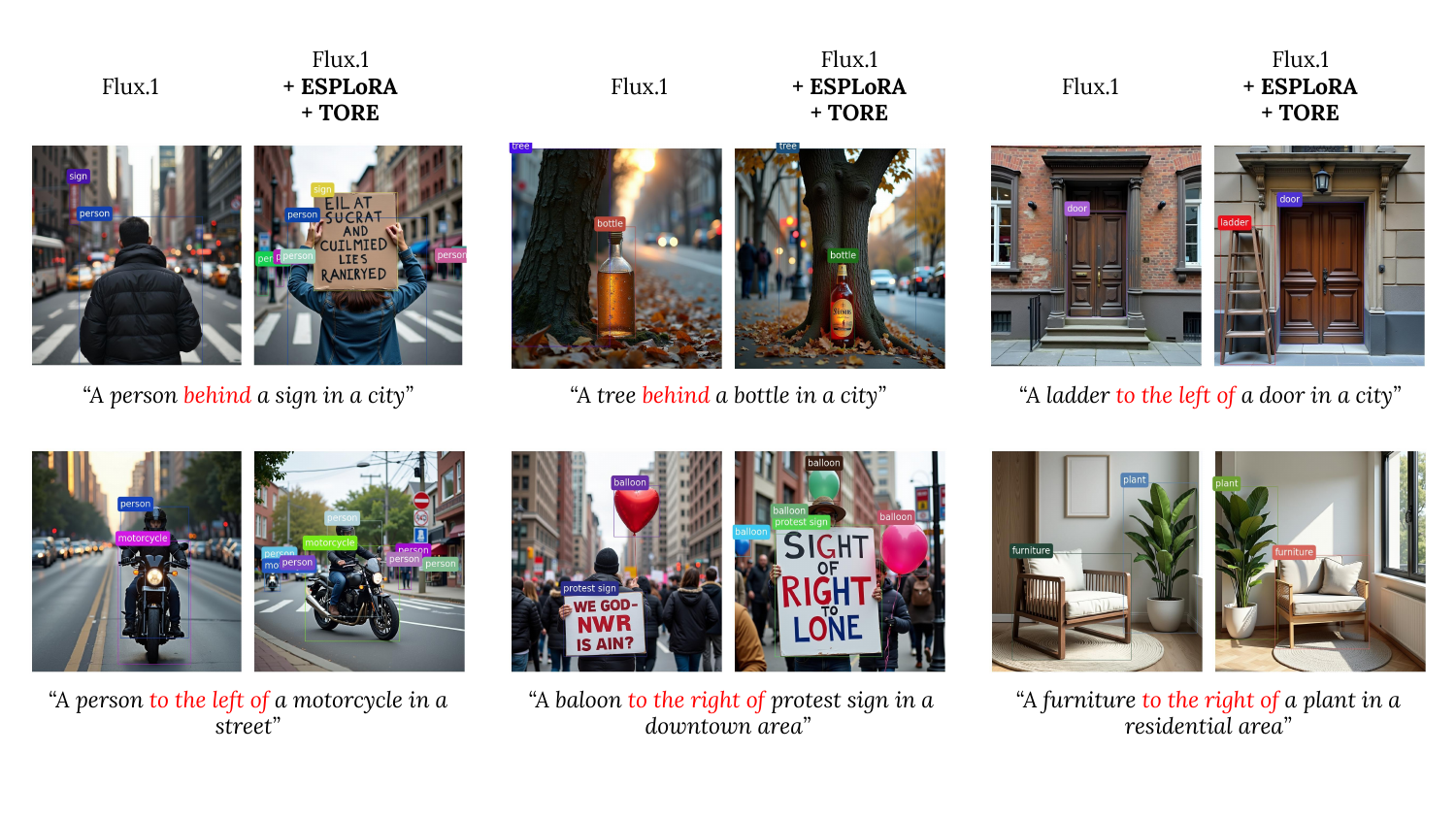}
    \caption{Comparison between Flux.1 and Flux.1 + ESPLoRA + TORE, samples from our Urban Benchmark, one relationship prompts.}
    \label{fig:qual-1r}
\end{figure*}

We present samples from our Urban Benchmark generated by Flux.1 + ESPLoRA + TORE compared with Flux.1 in \cref{fig:qual-1r} and additional qualitatives in non-urban scenarios in \cref{fig:extra-quals-flux}.
% We also report qualitatives for SDXL + ESPLoRA + TORE in \cref{fig:sdxl-quals}.
Our model displays improved spatial accuracy, while maintaining high quality in the generated content.
We also evaluate our model on the Non spatial spit of T2ICompBench and on textual variations on T2ICompBench 2D as explained in the Results section. We report Non spatial qualitatives and textual variations qualitatives for Flux.1 + ESPLoRA + TORE in \cref{fig:non-spatial-quals} and \cref{fig:variations-quals} respectively.
Our model is robust to alternative textual variations of relationships in prompts, and retains the general prompt following capabilities of the base Flux.1 model despite being fine-tuned on urban spatial prompts.
Finally, we compare our approach with CoMPaSS in a four-relationship scenario in \cref{fig:4-rel-comparison}. While both methods produce images that correctly capture the complex spatial relationships described in the prompt, our method demonstrates superior image quality.

\section{Dataset}
\begin{figure}
    \centering
    \begin{subfigure}[b]{0.23\linewidth}
        \includegraphics[width=\linewidth]{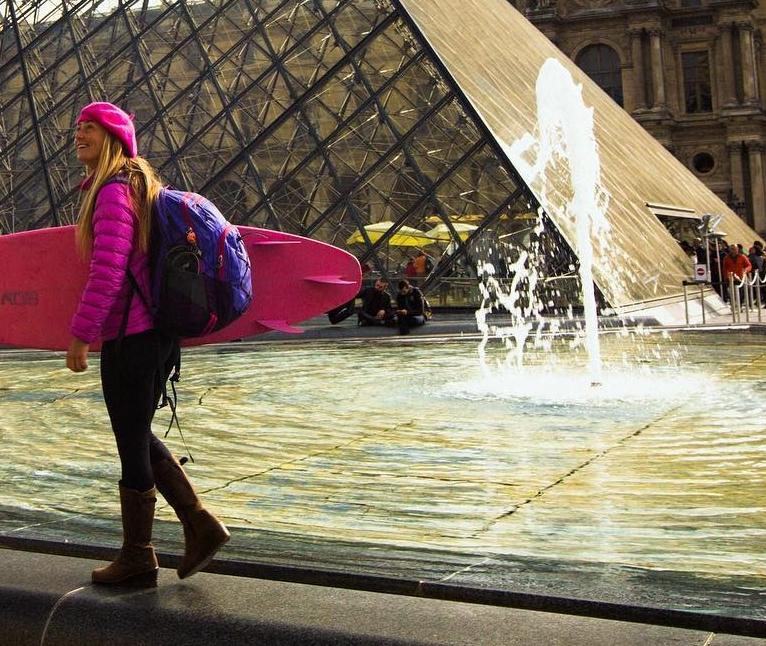}
        \caption{A fountain to the right of a person in a city}
    \end{subfigure}
    \begin{subfigure}[b]{0.23\linewidth}
        \includegraphics[width=\linewidth]{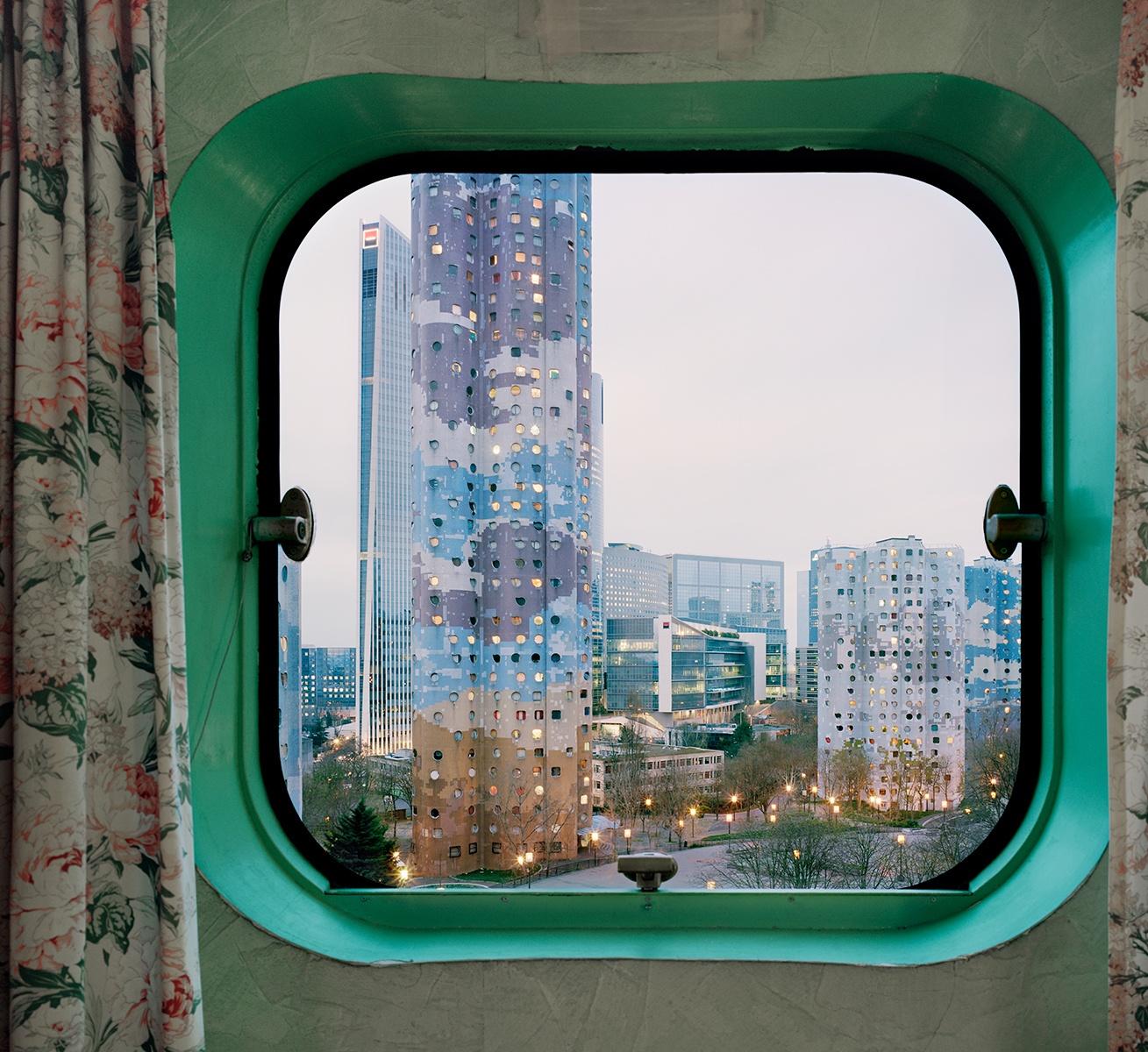}
        \caption{A window to the right of a curtain in a city}
    \end{subfigure}
    \vspace{0.3cm}
    \begin{subfigure}[b]{0.23\linewidth}
        \includegraphics[width=\linewidth]{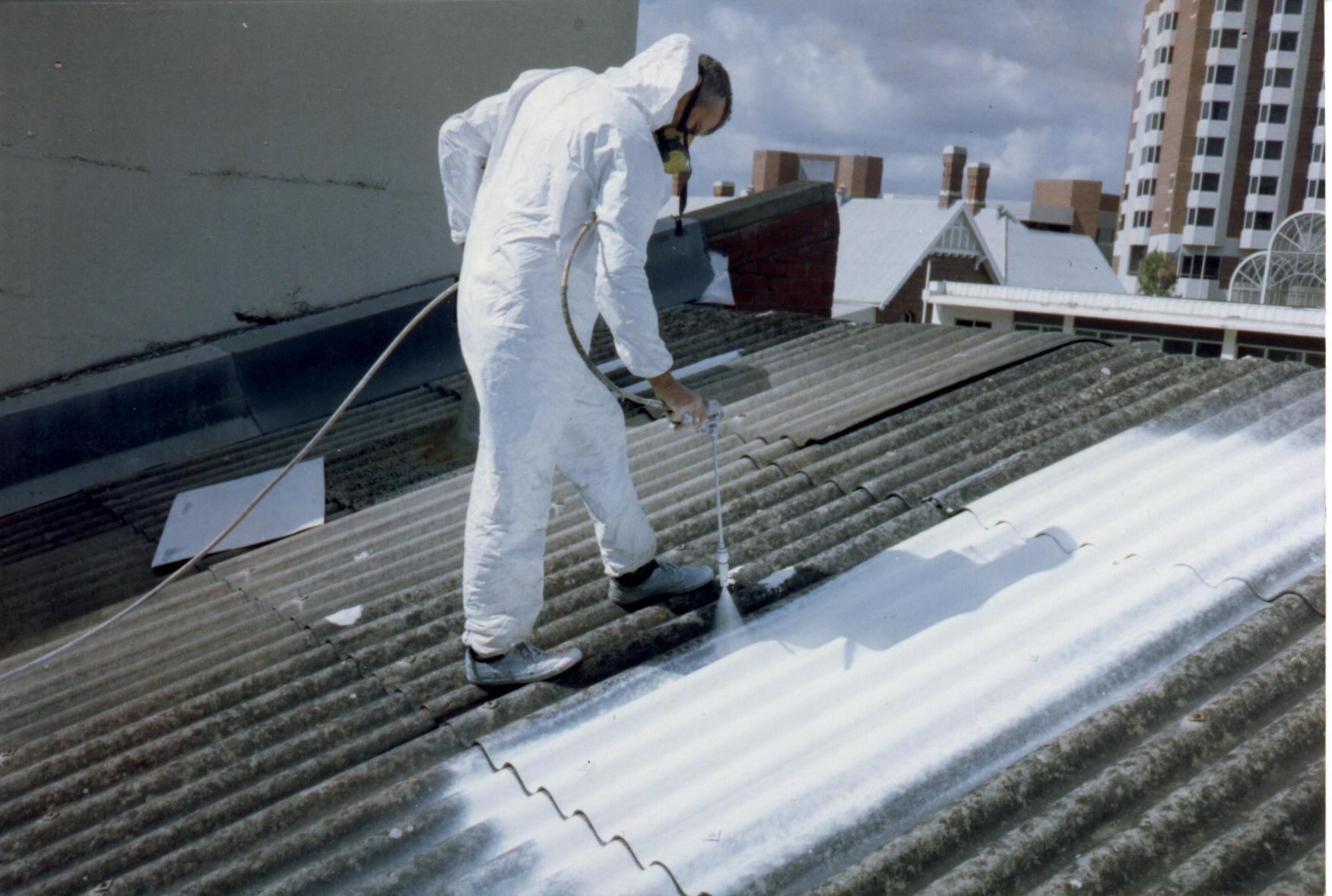}
        \caption{A rooftop under a high - rise, a cloud next to the high - rise in a city}
    \end{subfigure}
    \begin{subfigure}[b]{0.23\linewidth}
        \includegraphics[width=\linewidth]{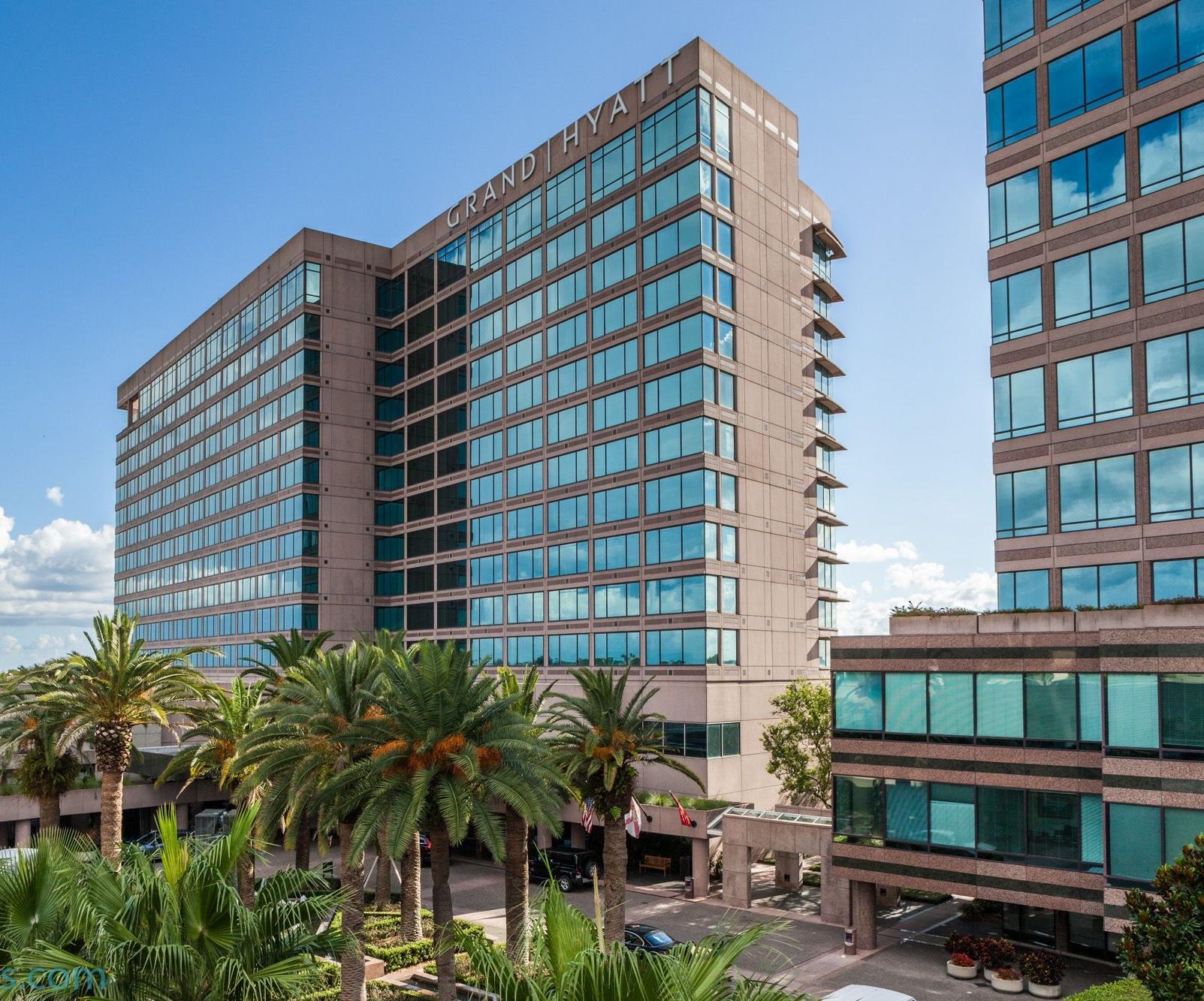}
        \caption{A palm tree next to a parking garage, a tall building to the left of the parking garage in a city}
    \end{subfigure}
    \caption{Examples of natural images with our procedurally generated prompts.}
    \label{fig:natural-examples}
\end{figure}

\begin{figure*}
    \centering
    \begin{subfigure}[b]{0.23\linewidth}
        \includegraphics[width=\linewidth]{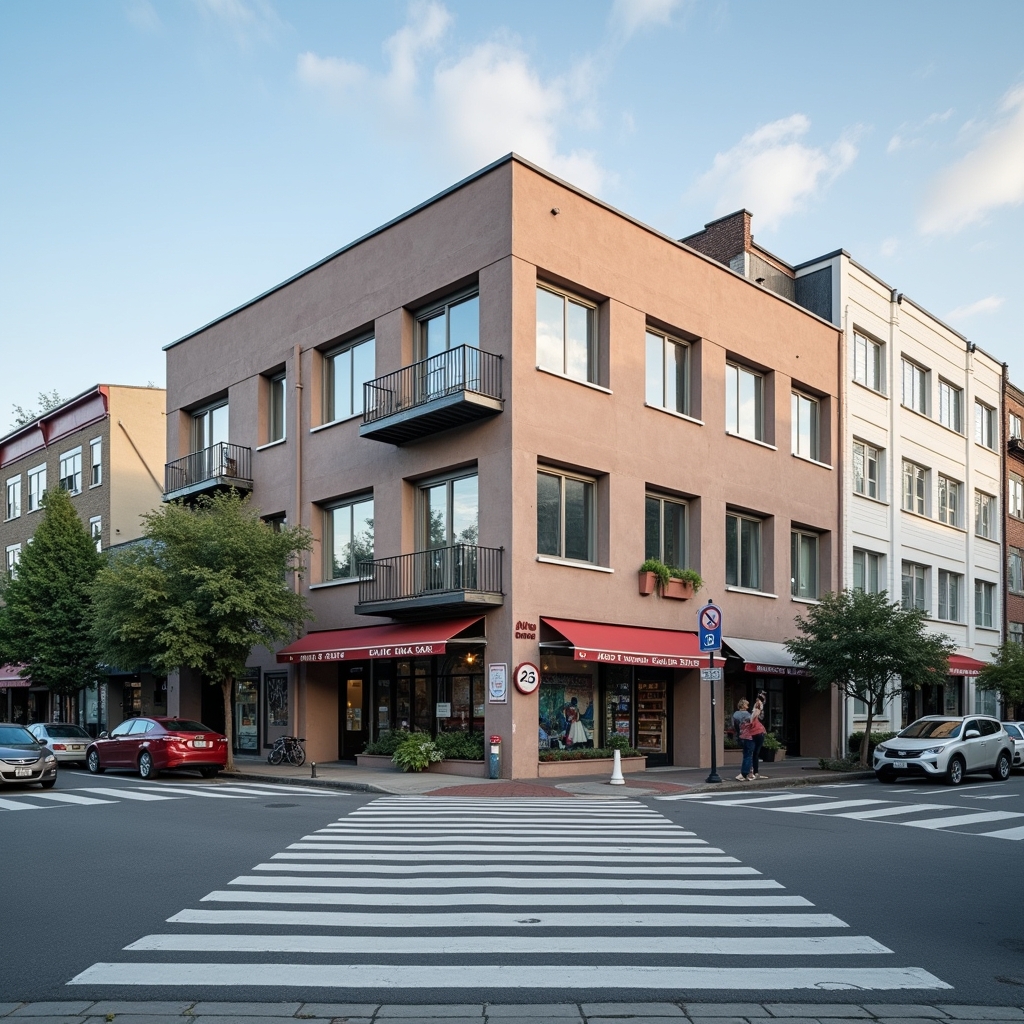}
        \caption{A building on top of a crosswalk in a city}
    \end{subfigure}
    \begin{subfigure}[b]{0.23\linewidth}
        \includegraphics[width=\linewidth]{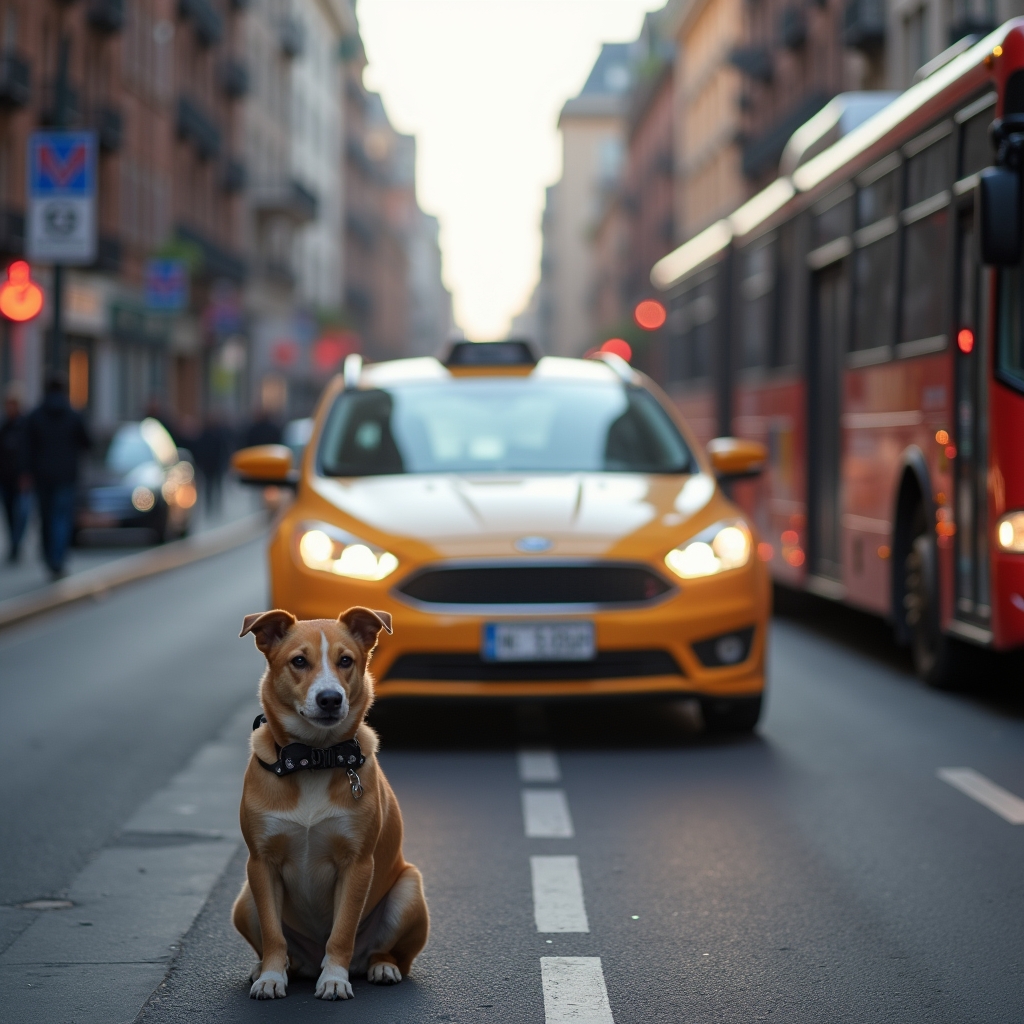}
        \caption{A bus to the right of a car in a city}
    \end{subfigure}
    \begin{subfigure}[b]{0.23\linewidth}
        \includegraphics[width=\linewidth]{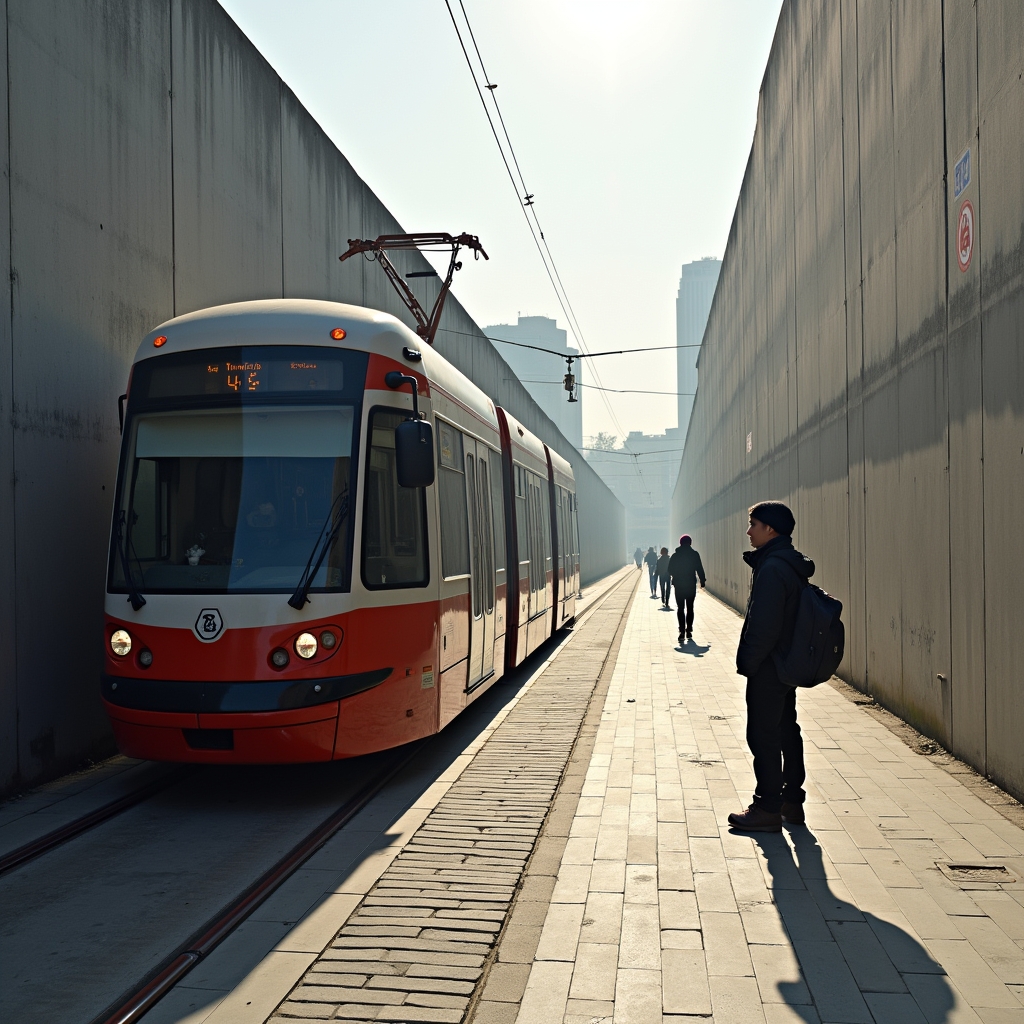}
        \caption{A tram to the left of a person, a concrete wall next to the person in a city}
    \end{subfigure}
    \begin{subfigure}[b]{0.23\linewidth}
        \includegraphics[width=\linewidth]{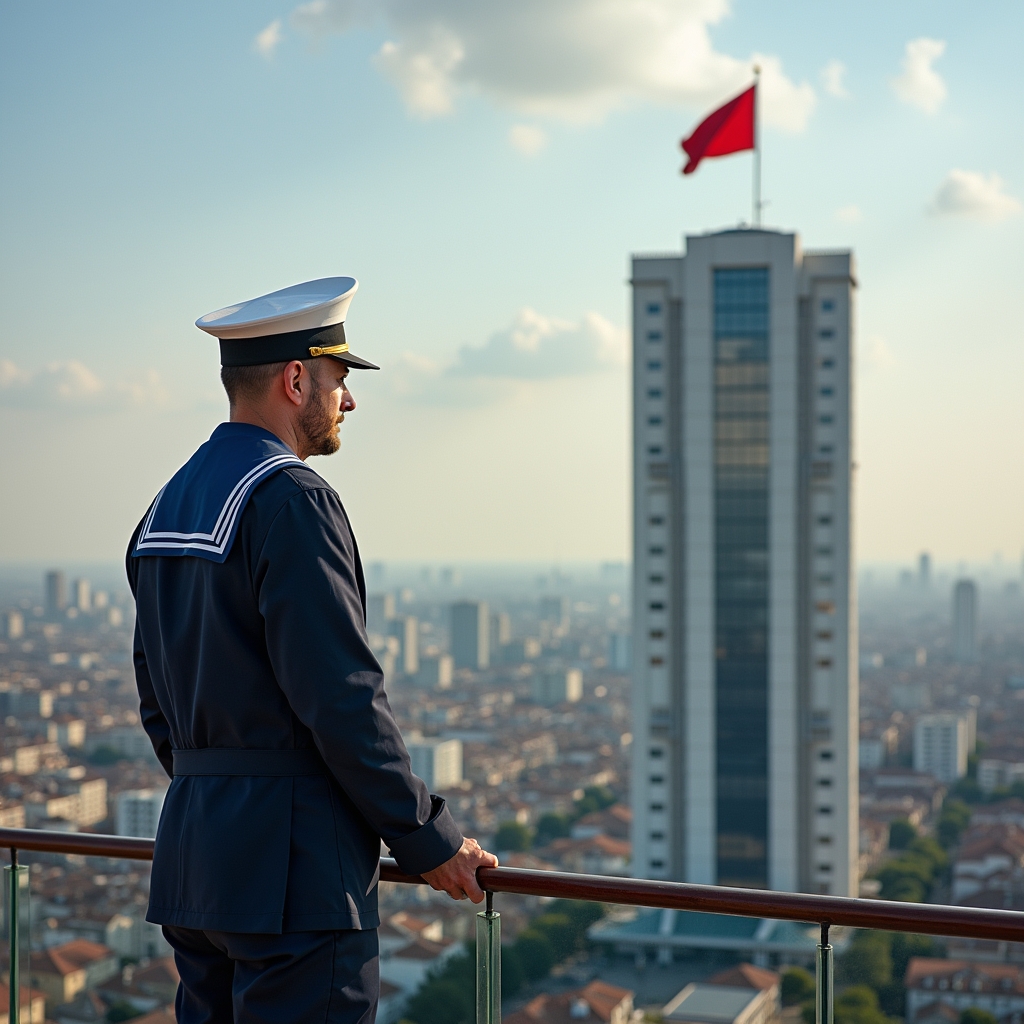}
        \caption{A sailor next to a building, a flag on top of the building in a city}
    \end{subfigure}
    \caption{Examples of synthetic images with our spatially accurate prompts.}
    \label{fig:synthetic-examples}
\end{figure*}

\cref{tab:synth-dataset-stats} shows the distribution of the synthetic dataset's prompts used for training.
We also provide some examples of our spatially accurate prompts in \cref{tab:example-prompts}, some natural samples in \cref{fig:natural-examples}, and some synthetic samples in \cref{fig:synthetic-examples}.
Our pipeline reliably captures spatial relationships in both natural and synthetic images, enabling the creation of a dataset with images paired to precisely aligned spatial prompts.

\section{Training Details}
We conduct all our experiments on state-of-the-art diffusion models SDXL and Flux.1.
We provide all LoRA training hyperparameters in \cref{tab:train-hyperparams}.

\section{Limitations}
\begin{figure}
    \centering
    \includegraphics[width=\linewidth]{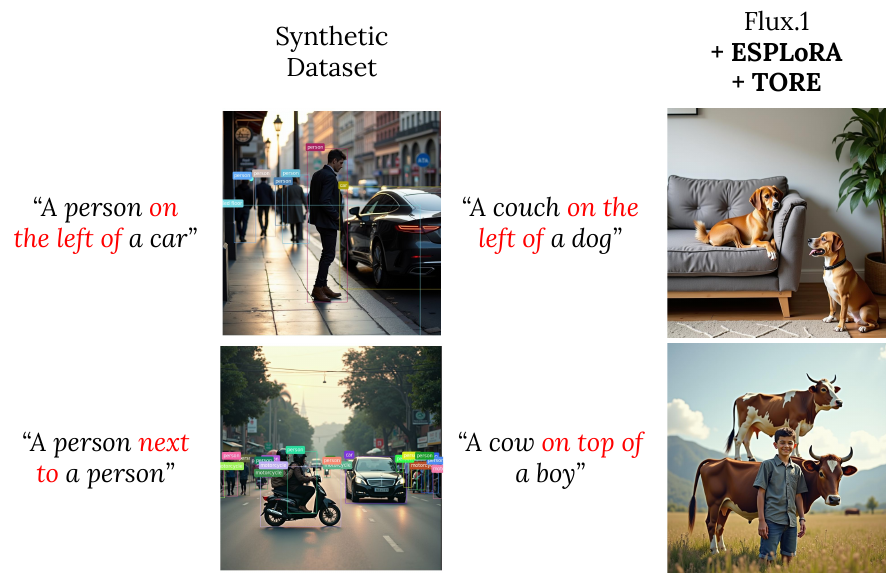}
    \caption{Samples showing a limitation of the ESPLoRA dataset construction pipeline. On the right, our model "cheats" by duplicating the requested subjects, increasing the likelihood of correctly matching the requested relationship.}
    \label{fig:limitations}
\end{figure}

A limitation of our dataset construction pipeline is that, in some cases (i.e. \cref{fig:limitations} left column), we capture a relation such as "a person next to a person", while the corresponding image actually contains many people. Although the relation is correct for at least one pair of objects, other instances in the image also satisfy the same relation. This sometimes leads the model to "cheat" by satisfying the required spatial relation through object duplication, like in \cref{fig:limitations} right column, thereby increasing the likelihood of matching the relation requested in the prompt.
Possible fixes include leveraging the depth map to enforce that the target objects appear in the foreground, while constraining them to be spatially close to each other and cropping out the rest of the image. In this way, the objects of interest remain at the center of attention, while other potentially valid instances are excluded.

\begin{figure}
    \centering
    \includegraphics[width=0.7\linewidth]{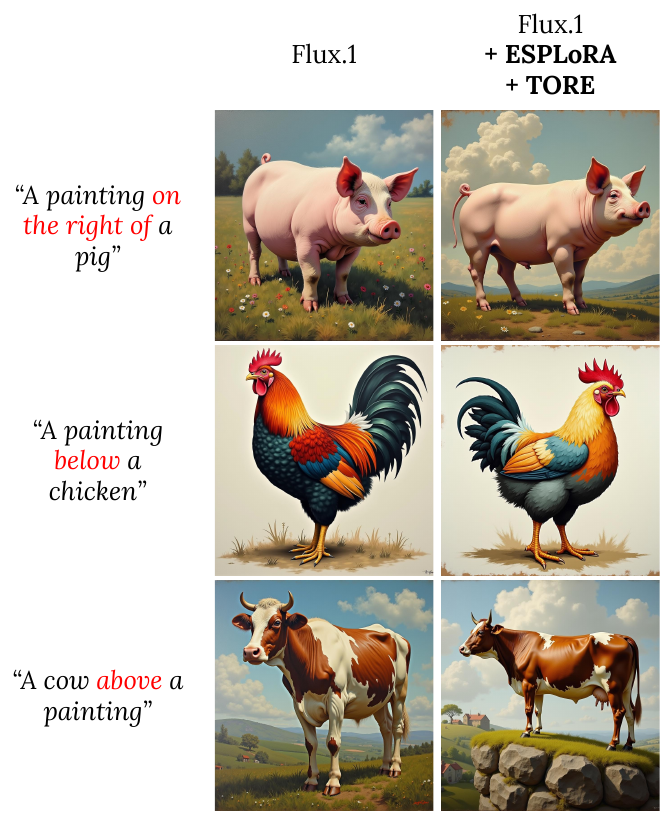}
    \caption{Example of bias in interpreting prompts. Both the base model and the fine-tuned model misinterpret prompts such as “a painting on the right of” as “a painting of,” applying the style to the entire image instead of generating the objects in the requested relationship.}
    \label{fig:painting-limitations}
\end{figure}

Another limitation of the method is that both the base model and our fine-tuned model exhibit a bias when processing prompts such as “a painting on the right of” or “a photo on the right of.” These are often misinterpreted as “a painting of” or “a photo of,” which leads the model to apply the style (painting or photo) to the entire image rather than generating the requested object positioned to the right, as shown in \cref{fig:painting-limitations}.

\section{Extracting relationships from bounding boxes}
\label{sec:apx-rel-metric}
T2I-CompBench introduces a geometric metric for extracting spatial relationships from bounding boxes during evaluation. It consists of a set of geometric constraints on bounding boxes center distances, each corresponding to a specific relationship type.
Given the relationship that the model is expected to generate, the appropriate constraint is applied.
While this might be good enough for evaluation, where the relationships that should appear on the image are known, we find it too ambiguous to extract relationships for training purposes.
In this setting, the relationship present in the image is unknown, requiring the metric to be applied iteratively across all possible relationships. As a result, we often observe that the T2I-CompBench metric assigns objects to multiple relationships simultaneously. We find this ambiguity undesirable in training data, as it could lead to unclear associations between prompts and spatial relationships in the model.

Therefore, we propose a stricter metric based solely on the boxes' coordinates.

Let \( B_1, B_2, B_3 \) be objects' bounding boxes with coordinates:

$B_i = (x_{\min}^i, y_{\min}^i, x_{\max}^i, y_{\max}^i)$ with $i \in \{1,2,3\}$,
width $\text{w}_i = x_{\max}^i - x_{\min}^i$, and height $\text{h}_i = y_{\max}^i - y_{\min}^i $.

First, we define location-independent distances:
\begin{equation}
    \begin{aligned}
        x_{\max\_dist} &= x_{\max}^{1} - x_{\max}^{2}, &
        x_{\min\_dist} &= x_{\min}^{1} - x_{\min}^{2} \\
        y_{\max\_dist} &= y_{\max}^{1} - y_{\max}^{2}, &
        y_{\min\_dist} &= y_{\min}^{1} - y_{\min}^{2}
    \end{aligned}
\end{equation}
where if $w_1 < w_2$ or $h_1 < h_2$ we swap $B_1$ and $B2$ in the horizontal or vertical distance computation respectively.
We use the distance between the respective $x_{max}$ or $y_{max}$ to constrain the bounding boxes' relative position on the axis opposite to \textit{locality}.
For instance, if \textit{locality} is horizontal, like right or left, we require each object to not be higher or lower than the other by a certain threshold, to avoid catching objects that are both in a horizontal relationship and a vertical one, like top or bottom.

We then define location-dependent distances:
\begin{equation}
    \begin{aligned}
        x_{\text{distance}} &= 
        \begin{cases} 
            x_{\min}^1 - x_{\max}^2, & \text{if \textit{locality} = right} \\
            x_{\min}^2 - x_{\max}^1, & \text{if \textit{locality} = left}
        \end{cases} \\
        y_{\text{distance}} &= 
        \begin{cases} 
            y_{\min}^1 - y_{\max}^2, & \text{if \textit{locality} = bottom} \\
            y_{\min}^2 - y_{\max}^1, & \text{if \textit{locality} = top}
        \end{cases}
    \end{aligned}
\end{equation}

The final positional constraints take the following form:
\begin{align}
    \text{if \textit{locality}} = \text{right or left}: \nonumber \\
    x_{\text{distance}} &> -\frac{\min(w_1, w_2)}{\tau}\label{eq:apx-right-pos-constr} \\
    y_{\max\_dist} &< \frac{\min(h_1, h_2)}{\tau}\label{eq:apx-ymax-constr} \\
    y_{\min\_dist} &> -\frac{\min(h_1, h_2)}{\tau}\label{eq:apx-ymin-constr}
\end{align}

\begin{align}
    \text{if \textit{locality}} = \text{top or bottom}: \nonumber \\
    y_{\text{distance}} &> -\frac{\min(h_1, h_2)}{\tau} \\
    x_{\max\_dist} &< \frac{\min(w_1, w_2)}{\tau}\label{eq:apx-max-constr} \\
    x_{\min\_dist} &> -\frac{\min(w_1, w_2)}{\tau}\label{eq:apx-xmin-constr}
\end{align}
Taking as example the \textit{right} locality again, i.e. "$B_1$ \textit{right} $B_2$", \cref{eq:apx-right-pos-constr} is valid only when the left border of object one ($x_{\min}^1$) is to the right of the right border of object two ($x_{\max}^2$). If that is not the case, the distance between the two, and thus their overlap, must be less than then the smaller width divided by $\tau$.
\cref{eq:apx-ymax-constr,eq:apx-ymin-constr} instead are only valid when the distance between the upper border (or lower for \cref{eq:apx-ymin-constr}) of both objects ($y_{\max}^{1}$ and $y_{\max}^{2}$) is less then the smaller height divided by $\tau$. In other words, object one cannot be higher or lower than object two for more than the smaller height divided by $\tau$.
Thus $\tau$ becomes an arbitrary threshold determining the metric strictness.

For the \textit{next} locality we verify that at least one of the \textit{right} and \textit{left} constraints are valid. For the \textit{between} locality instead, we check all possible triplets of bounding boxes, and verify that \textit{left} constraint is valid for $B_1$ and $B_2$ and the \textit{right} constraint is valid for $B_3$ and $B_2$.

For 3D relationships, we apply \cref{eq:apx-ymax-constr,eq:apx-ymin-constr,eq:apx-max-constr,eq:apx-xmin-constr} to ensure that there is enough overlap between the objects since they should be in front of one another.
Then we compute average depth $d_1$ for box $B_1$ and $d_2$ for box $B_2$.
If $d_1 > d_2$ then $B_1$ is in front of $B_2$, otherwise the other way around.

\begin{table}[h]
\centering
\begin{tabular}{@{}ccc@{}}
\toprule
\textbf{Relationship} &
  \textbf{\begin{tabular}[c]{@{}c@{}}1 Relationship\\ prompts\end{tabular}} &
  \textbf{\begin{tabular}[c]{@{}c@{}}2 Relationships\\ prompts\end{tabular}} \\ \midrule
    Right   & 1846 & 2044 \\
    Left    & 1781 & 2070 \\
    Top     & 2073 & 1803 \\
    Bottom  & 1793 & 1521 \\
    Next    & 2157 & 1857 \\
    Between & 400  & 1576 \\
    Front   & 1791 & 1388 \\
    Behind  & 1607 & 1415 \\
\bottomrule
\end{tabular}
\caption{Synthetic training dataset prompts distribution}
\label{tab:synth-dataset-stats}
\end{table}

\begin{table}
    \centering
    \setlength{\tabcolsep}{2pt}
    \begin{tabular}{lcc}
        \toprule
        \textbf{Hyperparameter} & \textbf{FLUX.1} & \textbf{SDXL} \\
        \midrule
        AdamW Learning Rate (LR) & 1e-4 & 1e-4 \\
        AdamW $\beta_1$ & 0.9 & 0.9 \\
        AdamW $\beta_2$ & 0.999 & 0.999 \\
        AdamW $\epsilon$ & 1e-8 & 1e-8 \\
        AdamW Weight Decay & 1e-2 & 1e-2 \\
        EMA Decay & 0.99 & 0.99 \\
        \midrule
        LR scheduler & Constant & Constant \\
        Training Epochs & up to 6 & up to 6 \\
        Batch Size & 2 & 4 \\
        Gradient Accumulation & 2 & 1 \\
        Training Resolution & 1024 $\times$ 1024 & 1024 $\times$ 1024 \\
        LoRA rank & 128 & 128 \\
        Trained Parameters & MMDiT & U-Net \\
        Prompt Dropout Probability & 10\% & 10\% \\
        Precision & Float16 & Float16 \\
        \bottomrule
    \end{tabular}
    \caption{Hyperparameters used during training.}
    \label{tab:train-hyperparams}
\end{table}

\clearpage

\begin{table*}[h]
\centering
\begin{tabular}{p{0.35\linewidth} | p{0.6\linewidth}}
\toprule
  \textbf{1 Relationship prompts} &
  \textbf{2 Relationship prompts} \\ \midrule
    A streetlight to the right of a garbage in a residential area & A street sign behind a person, a sunglass next to the person in a street \\
    A market behind a building in a city & A gate to the right of a garage door, a garage door between a garage door and a stair in a residential area \\
    A water tank on top of a street in a downtown area & A garden under a person, a traffic light on top of the person in a city \\
    A light fixture in front of a garage in a residential area & A streetlight between two streetlights, a bench to the right of the streetlight in a downtown area \\
    An umbrella to the right of a car in a street & A parking meter under a building, a pipe behind the building in a street \\
    A stone walkway under a window in a residential area & A window behind a building, a map on top of the building in a city \\
    A pool under a chair in a residential area & A tv in front of a building, a clock in front of the building in a city \\
    A white line under a building in a residential area & A street light between a tripod and a fire hydrant, a camera on top of the tripod in a street \\
    A couch in front of a building in a street & A fountain to the right of a statue, a fence under the statue in a city \\
    A waterway to the right of a car in a downtown area & A sidewalk to the left of a cell phone, a wifi symbol on top of the cell phone in a street \\
\bottomrule
\end{tabular}
\caption{A few examples of our spatially accurate prompts}
\label{tab:example-prompts}
\end{table*}

\end{document}